\documentclass[12pt]{article}

\usepackage[slantedGreek]{mathptmx}
\usepackage{mathptmx}
\usepackage[scaled=0.90]{helvet}
\usepackage{courier}
\normalfont

\usepackage{amsmath,amsthm}
\usepackage{amsfonts}
\usepackage{graphics}
\usepackage{graphicx}
\usepackage{color}
\usepackage[all]{xy}
\usepackage{natbib}

\DeclareGraphicsExtensions{.ps}

\usepackage[format=hang,font=small,labelfont=bf]{caption}

\numberwithin{equation}{section}

\newtheorem{theo}{Theorem}[section]
\newtheorem{lem}[theo]{Lemma}
\newtheorem{cor}[theo]{Corollary}
\newtheorem{pro}[theo]{Proposition}
\theoremstyle{definition}
\newtheorem{rem}{Remark}[section]

\renewcommand{\thefootnote}{\fnsymbol{footnote}}

\definecolor{blue}{rgb}{0,0,0}
\definecolor{red}{rgb}{0,0,0}

\begin{document}

\begin{center}

{\bf \Large \textcolor{blue}{Operator norm convergence of} \\
spectral clustering on level sets}\\

\vspace{1cm}

Bruno \textsc{Pelletier}\footnote{Department of Mathematics; IRMAR ---
  UMR CNRS 6625; Universit\'e Rennes II; Place du Recteur Henri Le
  Moal, CS 24307; 35043 Rennes Cedex; France;
  \texttt{bruno.pelletier@univ-rennes2.fr}} and Pierre
\textsc{Pudlo}\footnote{I3M: Institut de Math\'ematiques et de
  Mod\'elisation de Montpellier --- UMR CNRS 5149; Universit\'e
  Montpellier II, CC 051; Place Eug\`ene
  Bataillon; 34095 Montpellier Cedex 5, France;
  \texttt{pierre.pudlo@univ-montp2.fr}}
\\



\end{center} 

\vspace{0.25cm}

\begin{abstract} 
  \noindent {\rm Following \citet{hartigan}, a cluster is defined as a
    connected component of the $t$-level set of the
    underlying density, i.e., the set of points for which the density
    is greater than $t$. 
    A clustering algorithm which combines a density estimate with
    spectral clustering techniques is proposed.  Our algorithm is
    composed of two steps.  First, a nonparametric density estimate is
    used to extract the data points for which the estimated density
    takes a value greater than $t$.  Next, the extracted
    points are clustered based on the eigenvectors of a graph
    Laplacian matrix.  Under mild assumptions, we prove the almost
    sure convergence in operator norm of the empirical graph Laplacian
    operator associated with the algorithm.  Furthermore, we give the
    typical behavior of the representation of the dataset into the
    feature space, which establishes the
    strong consistency of our proposed algorithm.
    \\


  }

\end{abstract}

\vspace{.3cm}

\renewcommand{\thefootnote}{\arabic{footnote}}
 
\setcounter{footnote}{0}
\newpage
\section{Introduction}
\label{sec:introduction}
The aim of data clustering, or unsupervised classification, is to
partition a data set into several homogeneous groups relatively
separated one from each other with respect to a certain distance or
notion of similarity.  There exists an extensive literature on
clustering methods, and we refer the reader to
\cite{anderberg,hartigan,mclachlan}, Chapter 10 in \cite{duda}, and
Chapter 14 in \cite{hastie} for general materials on the subject.  In
particular, popular clustering algorithms, such as Gaussian mixture
models or k-means, have proved useful in a number of applications, yet
they suffer from some internal and computational limitations.  Indeed,
the parametric assumption at the core of mixture models may be too
stringent, while the standard k-means algorithm fails at identifying
complex shaped, possibly non-convex, clusters.
\\

The class of {\it spectral clustering} algorithms is presently
emerging as a promising alternative, showing improved performance over
classical clustering algorithms on several benchmark problems and
applications; see e.g., \cite{ng,vonluxburg2007}.  An overview of
spectral clustering algorithms may be found in \cite{vonluxburg2007},
and connections with kernel methods are exposed in \cite{filippone}.
The spectral clustering algorithm amounts at embedding the data into a
feature space by using the eigenvectors of the similarity matrix in
such a way that the clusters may be separated using simple rules,
e.g. a separation by hyperplanes.  The core component of the spectral
clustering algorithm is therefore the similarity matrix, or certain
normalizations of it, generally called graph Laplacian matrices; see
\cite{chung}.  Graph Laplacian matrices may be viewed as discrete
versions of bounded operators between functional spaces.  The study of
these operators has started out recently with the works by
\cite{belkin2004,belkin2005,coifman,nadler,koltchinskii,gine,Hein07}, among
others, and the convergence of the spectral clustering algorithm has
been established in \cite{vonluxburg2008}.
\\

The standard k-means clustering leads to
the optimal quantizer of the underlying distribution; see
\cite{macqueen,pollard81,linder}. However, determining what the limit
clustering obtained in \cite{vonluxburg2008} represents for the
distribution of the data remains largely an open question.  As a
matter of fact, there exists many definitions of a cluster; see
e.g., \cite{vonluxburgdavid} or \cite{garcia}.  Perhaps the most
intuitive and precise definition of a cluster is the one introduced by
\cite{hartigan}.  Suppose that the data is drawn from a probability
density $f$ on $\mathbb{R}^d$ and let $t$ be a positive number in the
range of $f$.  Then a cluster in the sense of \cite{hartigan} is a
connected component of the $t$-level set
$$\mathcal{L}(t) = \big\{x\in\mathbb{R}^d\,:\,f(x)\geq t\big\}.$$
This definition has several advantages.  First, it is
geometrically simple.
Second, it offers the possibility of filtering out possibly meaningless clusters by keeping only the observations falling in a region of high density, 
\textcolor{red}{This proves useful, for instance, in the situation where the data exhibits a cluster structure but is contaminated by a uniform background noise, as illustrated in our simulations in Section~4.}

In this context, the level $t$ should be considered as a
resolution level for the data analysis.  Several clustering algorithms
have been introduced building upon Hartigan's definition.
In \cite{cuevas2000,cuevas2001}, clustering is performed by estimating
the connected components of $\mathcal{L}(t)$; see also the work by
\cite{azzalini}.
Hartigan's definition is also used in \cite{bcp} to define an estimate of the
number of clusters.\\

In the present paper, the definition of a cluster given by
\cite{hartigan} is adopted, and we introduce a spectral clustering
algorithm on estimated level sets.  More precisely, given a random
sample $X_1,\dots,X_n$ drawn from a density $f$ on $\mathbb{R}^d$, our
proposed algorithm is composed of two operations.  In the first step,
given a positive number $t$, we extract the observations for which
$\hat{f}_n(X_i) \geq t$, where $\hat{f}_n$ is a nonparametric density
estimate of $f$ based on the sample $X_1,\dots,X_n$.  In the second
step, we perform a spectral clustering of the extracted points.  The
remaining data points are then left unlabeled.
\\


\textcolor{red}{Our proposal is to study the asymptotic behavior of this algorithm}.
As mentioned above, strong interest has recently been shown in
spectral clustering algorithms, and the major contribution to the
proof of the convergence of spectral clustering is certainly due to
\cite{vonluxburg2008}. In \cite{vonluxburg2008}, the graph Laplacian
matrix is associated with some random operator acting on the Banach
space of continuous functions. They prove the collectively compact
convergence of those operators towards a limit operator. Under mild
assumptions, we strengthen their results by establishing the almost
sure convergence in {\it operator norm}, but in a smaller Banach space
(Theorem~\ref{theo:converge}). 
\textcolor{blue}{This operator norm convergence is more amenable than
  the slightly weaker notion of convergence established in
  \cite{vonluxburg2008}. For instance, it is easy to check that the
  limit operator, and the graph Laplacian matrices used in the
  algorithm, are continuous in the scale parameter $h$.
}

\textcolor{blue}{
  We also derive the asymptotic representation of the
  dataset in the feature space 
  in Corollary~\ref{theo:strong}. This result implies that
  the proposed algorithm
  is strongly consistent and that, asymptotically,
  observations of $\mathcal{L}(t)$ are assigned to the same cluster if
  and only if they fall in the same connected component of the level set
  $\mathcal{L}(t)$.
}
\\

The paper is organized as follows.  In Section~\ref{sec:spectral}, we
introduce some notations and assumptions, as well as our proposed
algorithm.  Section~\ref{sec:main} contains our main results, namely
the convergence in operator norm of the random operators, and the
characterization of the dataset embedded into the feature space.
We provide a numerical example with a simulated dataset in
Section~\ref{sec:simulations}. 
Sections~\ref{section:convergence} and \ref{sec:strong} are devoted to the
proofs.  At the end of the paper, a technical result on the geometry
of level sets is stated in Appendix A, some useful results of
functional analysis are summarized in Appendix B, and the theoretical
properties of the limit operator are given in Appendix C.

\section{Spectral clustering algorithm}
\label{sec:spectral}
\subsection{Mathematical setting and assumptions}
Let $\{X_i\}_{i\ge 1}$ be a sequence of i.i.d. random vectors in
$\mathbb R^d$, with common probability measure $\mu$.
Suppose that
$\mu$ admits a density $f$ with respect to the Lebesgue measure on
$\mathbb{R}^d$.  The \textit{ $t$-level set} of $f$ is denoted by
$\mathcal{L}(t)$, i.e.,
$$\mathcal{L}(t) = \big\{x\in\mathbb{R}^d\,:\,f(x)\geq t\big\},$$
for all positive level $t$, and given $a\leq b$,
$\mathcal{L}_a^b$ denotes the set $\{x\in\mathbb{R}^d\,:\,a\leq
f(x)\leq b\}$.
The differentiation operator with respect to $x$ is denoted by $D_x$.
We assume that $f$ satisfies the following conditions.
\begin{quote}
  \textbf{Assumption 1.} \textit{(i)} $f$ is of class $\mathcal{C}^2$
  on $\mathbb{R}^d$; \textit{(ii)} $\|D_x f\|>0$ on the 
  set $\{x\in\mathbb{R}^d\,:\, f(x)=t\}$;
  \textit{(iii)} $f$, $D_xf$, and $D^2_xf$ are uniformly bounded on
  $\mathbb R^d$.
\end{quote}
Note that under Assumption~1, $\mathcal{L}(t)$ is compact whenever $t$
belongs to the interior of the range of $f$.  Moreover,
$\mathcal{L}(t)$ has a finite number $\ell$ of connected components
${\mathcal C}_j$, $j=1,\ldots ,\ell$.  For ease of notation, the
dependence of $\mathcal{C}_j$ on $t$ is omitted.  The minimal
distance between the connected components of $\mathcal{L}(t)$ is
denoted by $d_{min}$, i.e.,
\begin{equation}
\label{eq:dmin}
d_{min}= \inf_{i\neq j} {\rm dist}\big(\mathcal{C}_i,\mathcal{C}_j\big).
\end{equation}

Let $\widehat{f}_n$ be a consistent density estimate of $f$ based on the
random sample $X_1,\dots,X_n$.  The $t$-level set of $\widehat{f}_n$ is
denoted by $\mathcal{L}_n(t)$, i.e.,
$$\mathcal{L}_n(t) = \big\{x\in\mathbb{R}^d\,:\,\widehat{f}_n(x)\geq t\big\}.$$
Let $J(n)$ be the set of integers defined by
$$J(n) = \big\{j\in\{1,\dots,n\}:\,\widehat{f}_n(X_j) \geq t\big\}.$$
The cardinality  of $J(n)$ is denoted by $j(n)$.

Let $k:\mathbb{R}^d\to\mathbb{R}_+$ be a fixed function.  The unit
ball of $\mathbb{R}^d$ centered at the origin is denoted by $B$,
and the ball centered at $x\in\mathbb{R}^d$ and of radius $r$ is
denoted by $x+rB$.  We assume throughout that the function
$k$ satisfies the following set of conditions.
\begin{quote}
  \textbf{Assumption 2.}  \textit{(i)} $k$ is of class $\mathcal{C}^2$
  on $\mathbb{R}^d$; \textit{(ii)} the support of $k$ is $B$;
  \textit{(iii)} $k$ is uniformly bounded from below on $B/2$ by some
  positive number; and \textit{(iv)} $k(-x)=k(x)$ for all
  $x\in\mathbb{R}^d$.
\end{quote}
Let $h$ be a positive number. 
We
denote by $k_h:\mathbb{R}^d\to\mathbb{R}_+$ the map defined by
$k_h(u)=k({u}/{h})$.


\subsection{Algorithm}
\label{sub:algorithm}
The first ingredient of our algorithm is the
\textit{similarity matrix} $\mathbf{K}_{n,h}$ whose elements are given
by
$$\mathbf{K}_{n,h}(i,j)  =  k_h(X_j-X_i),$$
and where the integers $i$ and $j$ range over the random set $J(n)$.
Hence $\mathbf{K}_{n,h}$ is a random matrix indexed by
$J(n)\times J(n)$, whose values depend on the function $k_h$, and
on the observations $X_j$ lying in the estimated level set
$\mathcal{L}_n(t)$.
Next, we introduce the diagonal \textit{normalization matrix}
$\mathbf{D}_{n,h}$ whose diagonal entries are given by
$$\mathbf{D}_{n,h}(i,i) = \sum_{j\in J(n)}\mathbf{K}_{n,h}(i,j)
,\quad i\in J(n).
$$
Note that the diagonal elements of $\mathbf{D}_{n,h}$ are
positive.
\\

The spectral clustering algorithm is 
based on the matrix $\mathbf{Q}_{n,h}$ defined by
$$\mathbf{Q}_{n,h} = \mathbf{D}_{n,h}^{-1} \mathbf{K}_{n,h}.$$
Observe that $\mathbf{Q}_{n,h}$ is a random Markovian transition
matrix.
Note also that the (random) eigenvalues of $\mathbf{Q}_{n,h}$ are
real numbers and that $\mathbf{Q}_{n,h}$ is diagonalizable. Indeed the
matrix $\mathbf{Q}_{n,h}$ is conjugate to the symmetric matrix
$\mathbf{S}_{n,h}:=\mathbf{D}_{n,h}^{-1/2} \mathbf
K_{n,h}\mathbf{D}_{n,h}^{-1/2}$ since we may write
\[
\mathbf{Q}_{n,h} =  \mathbf{D}_{n,h}^{-1/2}\mathbf{S}_{n,h}
\mathbf{D}_{n,h}^{1/2}.
\]
Moreover, the inequality $\|\mathbf{Q}_{n,h}\|_\infty\le 1$ implies that the spectrum
$\sigma(\mathbf{Q}_{n,h})$ is a subset of $[-1;+1]$. Let $1=\lambda_{n,1}\ge
\lambda_{n,2} \ge\ldots \ge \lambda_{n, j(n)}\ge -1$ be the
eigenvalues of $\mathbf{Q}_{n,h}$, where in this enumeration, an
eigenvalue is repeated as many times as its multiplicity.

To implement the spectral clustering algorithm, the data points of the
partitioning problem are first embedded into $\mathbb R^\ell$ by using
the eigenvectors of $\mathbf{Q}_{n,h}$ associated with the $\ell$
largest eigenvalues, namely $\lambda_{n,1}$, $\lambda_{n,2}$, \ldots
$\lambda_{n,\ell}$.  More precisely, fix a collection $V_{n,1}$,
$V_{n,2}$, \ldots, $V_{n,\ell}$ of such eigenvectors with components
respectively given by $V_{n,k}=\{V_{n,k,j}\}_{j\in J(n)}$, for
$k=1,\dots,\ell$.  Then the $j^{\rm th}$ data point, for $j$ in
$J(n)$, is represented by the vector $\rho_n({X}_j)$ of $\mathbb
R^\ell$ defined by $\rho_n({X}_j):=\{V_{n,k,j}\}_{1\le k\le \ell}$.
At last, the embedded points are partitioned using a classical
clustering method, such as the k-means algorithm for instance.


\subsection{Functional operators associated with the matrices of the
  algorithm}
As exposed in the Introduction, some functional operators are
associated with the matrices acting on $\mathbb{C}^{J(n)}$ defined in
the previous paragraph.  The link between matrices and functional
operators is provided by the evaluation map defined in
\eqref{eq:evalmap} below.  As a consequence, asymptotic results on the
clustering algorithm may be derived by studying first the limit
behavior of these operators.

To this aim, let us first introduce some additional notation.
For $\mathcal{D}$ a subset of $\mathbb{R}^d$, let $W(\mathcal{D})$ be
the Banach space of complex-valued, bounded, and continuously
differentiable functions with bounded gradient, endowed with the norm
$$\|g\|_{W} = \|g\|_\infty + \|D_x g\|_\infty.$$

Consider the 
non-oriented graph whose vertices are the $X_j$'s
for $j$ ranging in $J(n)$.  The similarity matrix $\mathbf{K}_{n,h}$
gives random weights to the edges of the graph and the random
transition matrix $\mathbf{Q}_{n,h}$ defines a random walk
on the vertices of a random graph.  Associated
with this random walk is the transition operator
$Q_{n,h}:W\big(\mathcal{L}_n(t)\big) \to W\big(\mathcal{L}_n(t)\big)$
defined for any function $g$ by
$$ Q_{n,h}g (x) =  
\int_{\mathcal{L}_n(t)} q_{n,h}(x,y)g(y)\mathbb{P}_n^t(dy).$$ In this
equation, $\mathbb{P}_n^t$ is the discrete random probability measure
given by
\[
\mathbb{P}_n^t = \frac{1}{j(n)} \sum_{j\in J(n)}\delta_{X_j},
\]
and
\begin{equation}
q_{n,h}(x,y) = \frac{k_h(y-x)}{K_{n,h}(x)},
\quad \mathrm{where}~
K_{n,h}(x) = \int_{\mathcal{L}_n(t)}k_h(y-x)\mathbb{P}_n^t(dy).
\label{eq:qnh}
\end{equation}
In the definition of $q_{n,h}$, we use the convention that $0/0=0$,
but this situation does not occur in the proofs of our results.\\

Given the \textit{evaluation map} $\pi_n :
{W}\big(\mathcal{L}_n(t)\big) \to \mathbb{C}^{J(n)}$
defined by
\begin{equation}
\label{eq:evalmap}
\pi_n(g) = \Big\{ g(X_j)\,:\, j\in J(n) \Big\},
\end{equation}
the matrix $\mathbf{Q}_{n,h}$ and the operator $Q_{n,h}$ are related by
$\mathbf{Q}_{n,h}\circ \pi_n = \pi_n\circ Q_{n,h}$.
Using this relation, asymptotic properties of the spectral
clustering algorithm may be deduced from the limit behavior of the sequence of
operators $\{Q_{n,h}\}_n$.  The difficulty, though, is that $Q_{n,h}$
acts on ${W}\big(\mathcal{L}_n(t)\big)$ and $\mathcal{L}_n(t)$
is a random set which varies with the sample.  For this reason, we
introduce a sequence of operators $\widehat{Q}_{n,h}$ acting on
$W\big(\mathcal{L}(t)\big)$ and constructed from $Q_{n,h}$ as follows.\\

First of all, recall that under Assumption~1, the gradient of $f$ does
not vanish on the set $\{x\in\mathbb{R}^d\,:\,f(x)=t\}$.  Since $f$ is
of class $\mathcal{C}^2$, a continuity argument implies that there
exists $\varepsilon_0>0$ 
such that $\mathcal{L}_{t-\varepsilon_0}^{t+\varepsilon_0}$ contains
no critical points of $f$.
Under this condition, Lemma~\ref{lem:levelsets} states that
$\mathcal{L}(t+\varepsilon)$ is diffeomorphic to $\mathcal{L}(t)$ for
every $\varepsilon$ such that $|\varepsilon|\leq\varepsilon_0$. 
In all of the following, it is assumed that $\varepsilon_0$ is small
enough so that
\begin{equation}
  \label{eq:vareps0}
  \varepsilon_0/\alpha(\varepsilon_0)<h/2, 
  \quad \text{where}~
  \alpha(\varepsilon_0)=\inf\big\{\|D_xf(x)\|;\,
  x\in\mathcal{L}_{t-\varepsilon_0}^t\big\}.
\end{equation}
Let
$\{\varepsilon_n\}_n$ be a sequence of positive numbers such that
$\varepsilon_n\leq\varepsilon_0$ for each $n$, and $\varepsilon_n\to0$ as
$n\to\infty$.  In Lemma~\ref{lem:levelsets} an explicit
diffeomorphism $\varphi_n$  carrying 
$\mathcal{L}(t)$ to $\mathcal{L}(t-\varepsilon_n)$ is constructed,
i.e.,
\begin{equation}
\label{eq:phin}
\varphi_n:\mathcal{L}(t) \stackrel{\cong}{\longrightarrow} 
\mathcal{L}(t-\varepsilon_n).
\end{equation}
The diffeomorphism $\varphi_n$ induces the linear operator
$\Phi_n:W\big(\mathcal{L}(t)\big)\to
W\big(\mathcal{L}(t-\varepsilon_n)\big)$ defined by 
$\Phi_n g = g\circ\varphi_n^{-1}.$\\

Second, let $\Omega_n$ be the probability event defined by
\begin{equation}
\label{eq:omega_n}
\Omega_n = \Big[\|\widehat{f}_n-f\|_\infty \leq \varepsilon_n \Big] 
\cap 
\left[\inf \left\{\|D_x\widehat{f}_n(x)\|,
x\in {\mathcal{L}_{t-\varepsilon_0}^{t+\varepsilon_0}}\right\}
\geq \frac{1}{2}\|D_x f\|_\infty\right].
\end{equation}
Note that on the event $\Omega_n$, the following inclusions hold:
\begin{equation}
\label{eq:inclusions}
\mathcal{L}(t-\varepsilon_n)
\subset
\mathcal{L}_n(t)\subset\mathcal{L}(t+\varepsilon_n).
\end{equation}
We assume that the indicator function $\mathbf 1_{\Omega_n}$
tends to $1$ almost surely as $n\to\infty$, which is satisfied by
common density estimates $\widehat{f}_n$ under mild assumptions.  For
instance, consider a kernel density estimate with a Gaussian kernel.
Then for a density $f$ satisfying the conditions in Assumption~1, we
have $\|D_x^{(p)}\widehat{f}_n - D_x^{(p)}f\|_\infty \to 0$ almost
surely as $n\to\infty$, for $p=0$ and $p=1$ (see e.g.,
\cite{prakasa}), which implies that $\mathbf{1}_{\Omega_n} \to 1$
almost surely as $n\to\infty$.
\\

We are now in a position to introduce the operator $\widehat{Q}_{n,h}:
W\big(\mathcal{L}(t)\big) \to
W\big(\mathcal{L}(t)\big)$ defined on the event $\Omega_n$
by
\begin{equation}
\label{eq:qqhat}
\widehat{Q}_{n,h} = \Phi_n^{-1}Q_{n,h}\Phi_n,
\end{equation}
and we extend the definition of $\widehat{Q}_{n,h}$ to the whole
probability space by setting it to the null operator on the complement
$\Omega_n^c$ of $\Omega_n$. In other words, on $\Omega_n^c$, the function
$\widehat{Q}_{n,h}g$ is identically zero for each $g\in
W\big(\mathcal{L}(t)\big)$. 

\begin{rem}
  \label{rem:omega_n}
  Albeit the relevant part of $\widehat{Q}_{n,h}$ is defined
  on $\Omega_n$ for technical reasons, this does not bring any
  difficulty as long as one is concerned with almost sure convergence.
  To see this, let $(\Omega,\mathcal A, P)$ be the probability
  space on which the $X_i$'s are defined.  Denote by $\Omega_\infty$
  the event on which $\mathbf 1_{\Omega_n}$ tends to 1, and recall
  that $P(\Omega_\infty)=1$ by assumption. Thus, for every
  $\omega\in\Omega$, there exists a random integer $n_0(\omega)$ such
  that, for each $n\ge n_0(\omega)$, $\omega$ lies in
  $\Omega_n$. Besides $n_0(\omega)$ is finite on $\Omega_\infty$.
  Hence in particular, if $\{Z_n\}$ is a sequence of random variables
  such that $Z_n\mathbf 1_{\Omega_n}$ converges almost surely to some
  random variable $Z_\infty$, then $Z_n\to Z_\infty$ almost surely.
\end{rem}

\section{Main results}
\label{sec:main}

Our main result (Theorem~\ref{theo:converge}) states that
$\widehat{Q}_{n,h}$ converges in operator norm to the limit operator
$Q_h:W\big(\mathcal{L}(t)\big) \to W\big(\mathcal{L}(t)\big)$ defined
by
\begin{equation}
Q_hg (x) = \int_{\mathcal{L}(t)} q_h(x,y)g(y)\mu^t(dy),\label{eq:Qh}
\end{equation}
where $\mu^t$ denotes the conditional distribution of $X$ given the
event $\big[X\in\mathcal{L}(t)\big]$, and where
\begin{equation}
  q_h(x,y) = \frac{k_h(y-x)}{K_h(x)},
  \quad
  \mathrm{with}~
  K_h(x) = \int_{\mathcal{L}(t)} k_h(y-x)\mu^t(dy). \label{eq:qh}
\end{equation}

\begin{theo}[Operator Norm Convergence]
\label{theo:converge}
\textcolor{blue}{Suppose that Assumptions 1 and 2 hold.} We have
\[\big\|\widehat{Q}_{n,h} - Q_h\big\|_W \to 0\quad
\text{almost surely as }n\to\infty.\]
\end{theo}
The proof of Theorem~\ref{theo:converge} is given in
Paragraph~\ref{sub:convergence}.
\textcolor{red}{Its main arguments are as follows. First, the three
  classes of functions defined in Lemma~\ref{lem:gc} are shown to be
  Glivenko-Cantelli. This, together with additional technical results,
  leads to uniform convergences of some linear operators
  (Lemma~\ref{lem:rconv}). } 

\textcolor{red}{Theorem~\ref{theo:converge} implies the consistency of
  our algorithm. We recall that $d_{min}$ given in \eqref{eq:dmin} is
  the minimal distance between the connected components of the level
  set.
  The starting point is the fact that, provided that
  $h<d_\text{min}$, the connected components of the level set
  $\mathcal L(t)$ are the recurrent classes of the Markov chain whose
  transitions are given by $Q_h$.  Indeed, this process cannot jump
  from one component to the other ones. Hence, $Q_h$ defines the
  desired clustering via its eigenspace corresponding to the
  eigenvalue $1$.  }


\textcolor{blue}{ As stated in Proposition~\ref{pro:harmonic} in the
  Appendices, the eigenspace of the limit operator $Q_h$ associated
  with the eigenvalue $1$ is spanned by the indicator functions of the
  connected components of $\mathcal{L}(t)$. Hence the representation
  of the extracted part of the dataset into the feature space $\mathbb
  R^\ell$ (see the end of Paragraph~\ref{sub:algorithm}) tends to
  concentrate around $\ell$ different centroids. Moreover, each of these
  centroids corresponds to a cluster, i.e., to a connected component
  of $\mathcal{L}(t)$.  }

\textcolor{blue}{ More precisely, using the convergence in operator
  norm of $\widehat{Q}_{n,h}$ towards $Q_h$, together with the results
  of functional analysis given in Appendix \ref{appendix:B}, we obtain
  the following corollary which describes the asymptotic behavior of
  our algorithm.  Let us denote by $J(\infty)$ the set of 
  integers $j$ such that $X_j$ is in the level set $\mathcal
  L(t)$. For all $j\in J(\infty)$, define $k(j)$ as the integer such
  that $X_j\in \mathcal C_{k(j)}$.  }
\begin{cor}
  \label{theo:strong}
  \textcolor{blue}{Suppose that Assumptions 1 and 2 hold, and that
    $h$ is in $(0;d_{min})$.} 
  There exists a sequence $\{\xi_n\}_n$ of linear transformations
  of $\mathbb R^\ell$ such that, for all $j\in J(\infty)$, $\xi_n
  \rho_n(X_j)$ converges almost surely to $e_{k(j)}$, where $e_{k(j)}$
  is the vector of $\mathbb R^\ell$
 \textcolor{red}{whose components are all 0 except the $k(j)^{\rm th}$ component equal to $1$.}
\end{cor}


Corollary~\ref{theo:strong}, which is new up to our knowledge, is proved in
  Section~\ref{sec:strong}.
Corollary~\ref{theo:strong} states that the data points embedded in the feature space concentrate on separated centroids.
   As a consequence, any partitioning algorithm
  (e.g., $k$-means) applied in the feature space will asymptotically
  yield the desired clustering. In other words, the clustering
  algorithm is consistent.  Note that if one is only interested in the
  consistency property, then this result could be obtained through
  another route.  Indeed, it is shown in \cite{bcp} that the
  neighborhood graph with connectivity radius $h$ has asymptotically
  the same number of connected components as the level set.  Hence,
  splitting the graph into its connected components leads to the
  desired clustering as well. 
But Corollary~\ref{theo:strong}, by giving the asymptotic representation of the data when embedded in the feature space $\mathbb{R}^\ell$, provides additional insight into spectral clustering algorithms.
In particular, Corollary~\ref{theo:strong} provides a rationale for the heuristic of \cite{zelnik-manor} for automatic selection of the number of groups.
Their idea is to quantify the amount of concentration of the points embedded in the feature space, and to select the number of groups leading to the maximal concentration.
Their method compared favorably with the eigengap heuristic considered in \cite{vonluxburg2007}.\\

Naturally, the selection of the number of groups is also linked with the choice of the parameter $h$.
In this direction, let us emphasize that the operators $\widehat{Q}_{n,h}$ and $Q_h$ depend continuously on
the scale parameter $h$. Thus, the spectral properties of both
operators will be close to the ones stated in Corollary~\ref{theo:strong},
if $h$ is in the neighborhood of the interval $(0;d_{min})$. This follows from the continuity of an isolated set of eigenvalues, as stated
in Appendix~\ref{appendix:B}.  In particular, the sum of the eigenspaces
of $Q_h$ associated with the eigenvalues close to $1$ is spanned by
functions that are close to (in $W(\mathcal L(t))$-norm) the
indicator functions of the connected components of $\mathcal
L(t)$. Hence, the representation of the dataset in the feature space
$\mathbb R^\ell$ still concentrates on some neighborhoods of $e_k$,
$1\le k \le \ell$ and a simple clustering algorithm such as $k$-means
will still give the desired result.  To sum up the above, if
assumptions 1 and 2 hold, our algorithm is consistent for all $h$ in
$(0,h_{max})$ for some $h_{max}>d_{min}$.\\

Several questions, though, remain largely open.
  For instance, one might ask if a similar result holds for the
  classical spectral clustering algorithm, i.e., without the
  preprocessing step.  This case corresponds to taking $t=0$.  One
  possibility may then be to consider a sequence $h_n$, with $\lim
  h_n=0$ and to the study the limit of the operator $Q_{n,h_n}$.



\section{Simulations}
\label{sec:simulations}

\begin{figure}[htb]
\begin{center}
\begin{tabular}{cc}
  \includegraphics[width=0.5\linewidth]{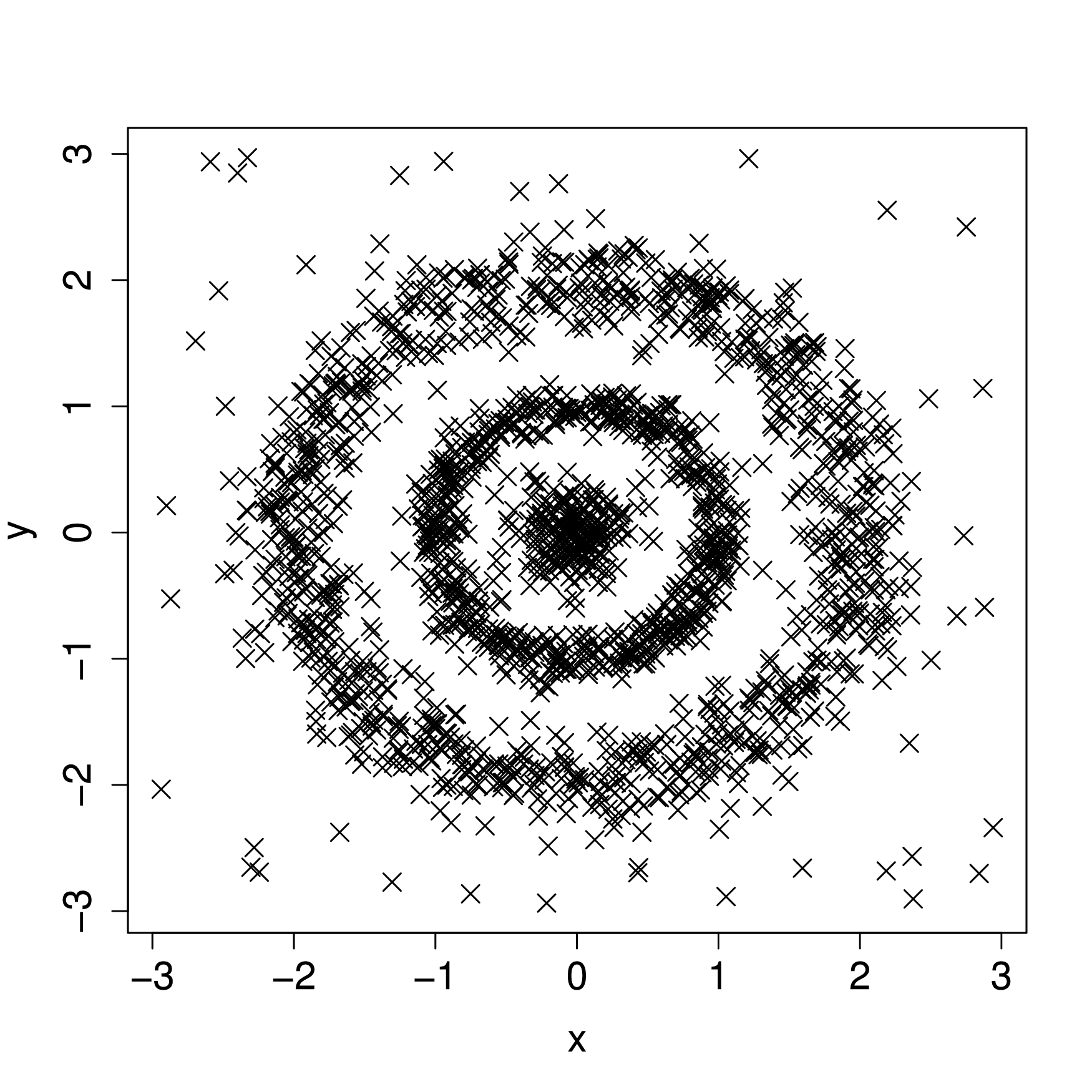} & 
  \includegraphics[width=0.5\linewidth]{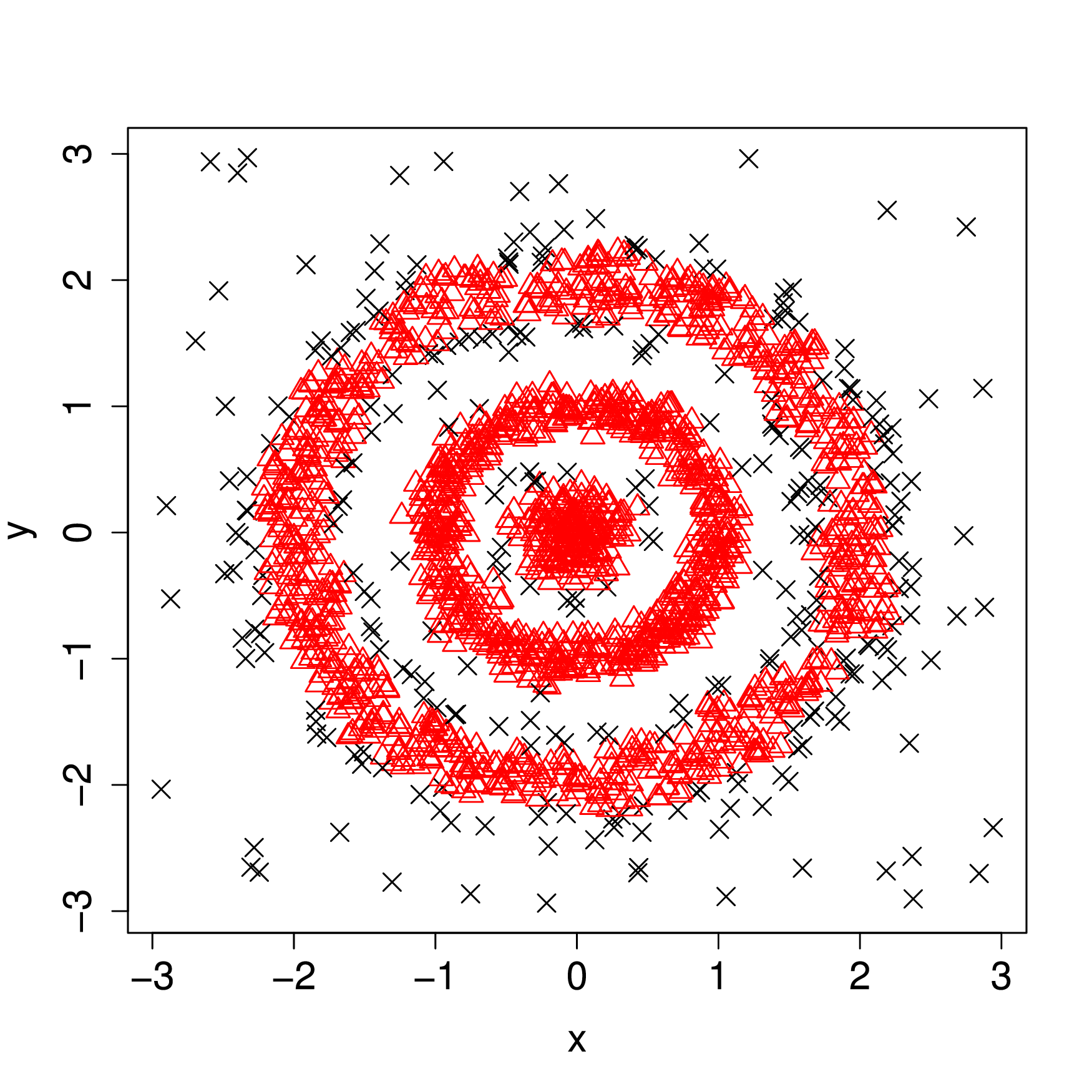}
\end{tabular}
\end{center}
\caption{{\it Left}: simulated points. {\it Right}: Points belonging
  to the estimated level set (red triangle) and remaining points (dark cross).}
\label{fig:points}
\end{figure}

\begin{figure}
\begin{center}
\begin{tabular}{cc}
  \includegraphics[width=0.5\linewidth]{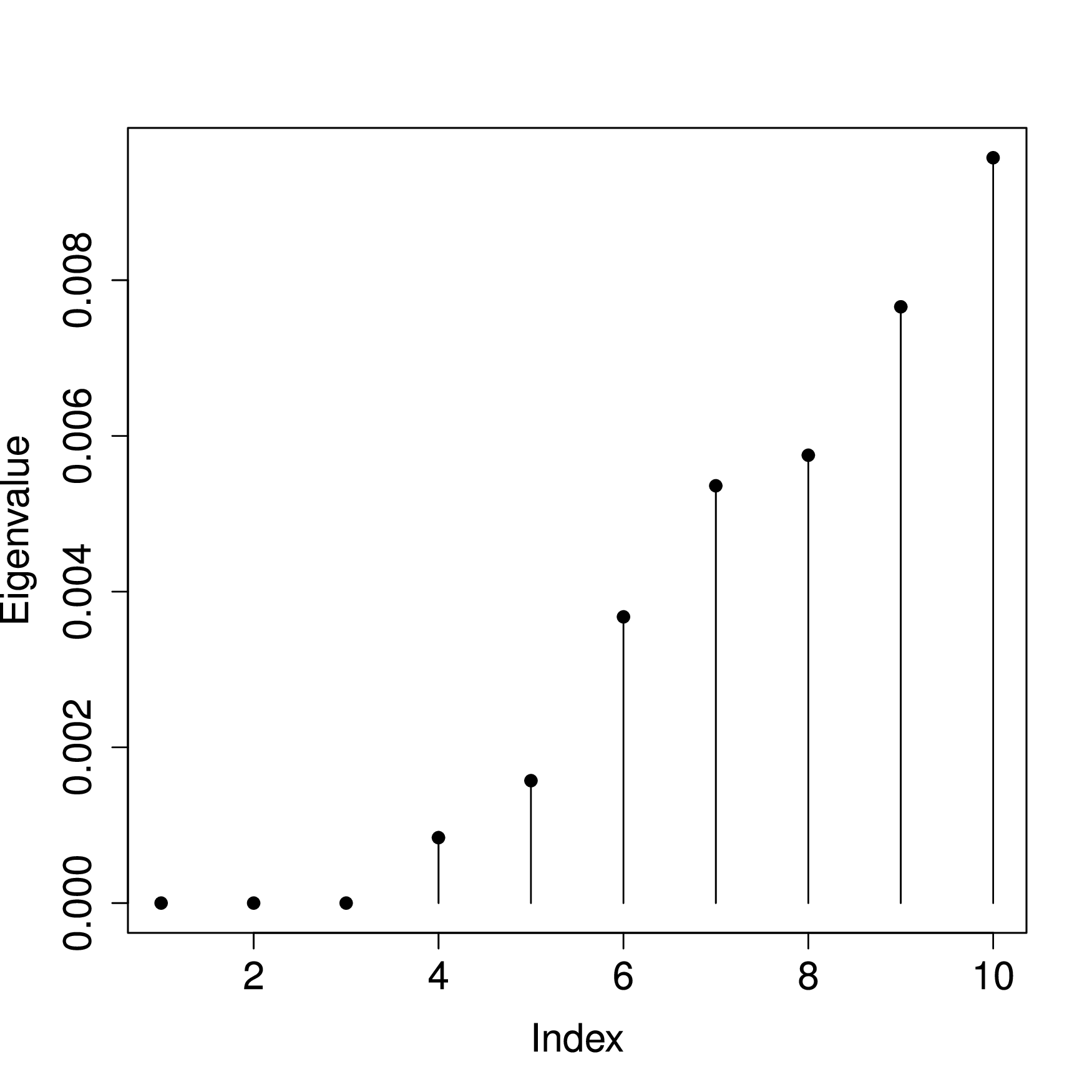} 
  & \includegraphics[width=0.5\linewidth]{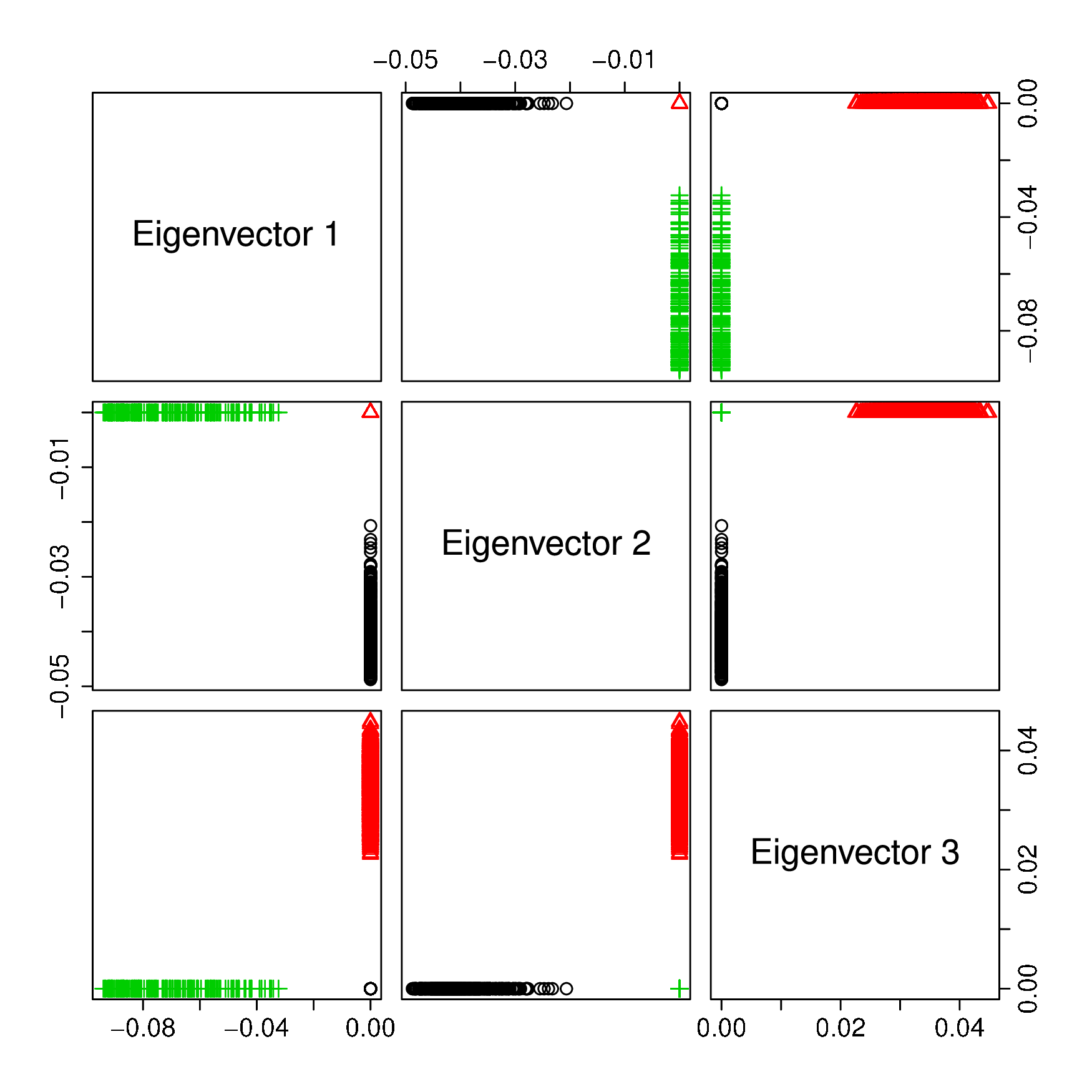}\\
  \includegraphics[width=0.5\linewidth]{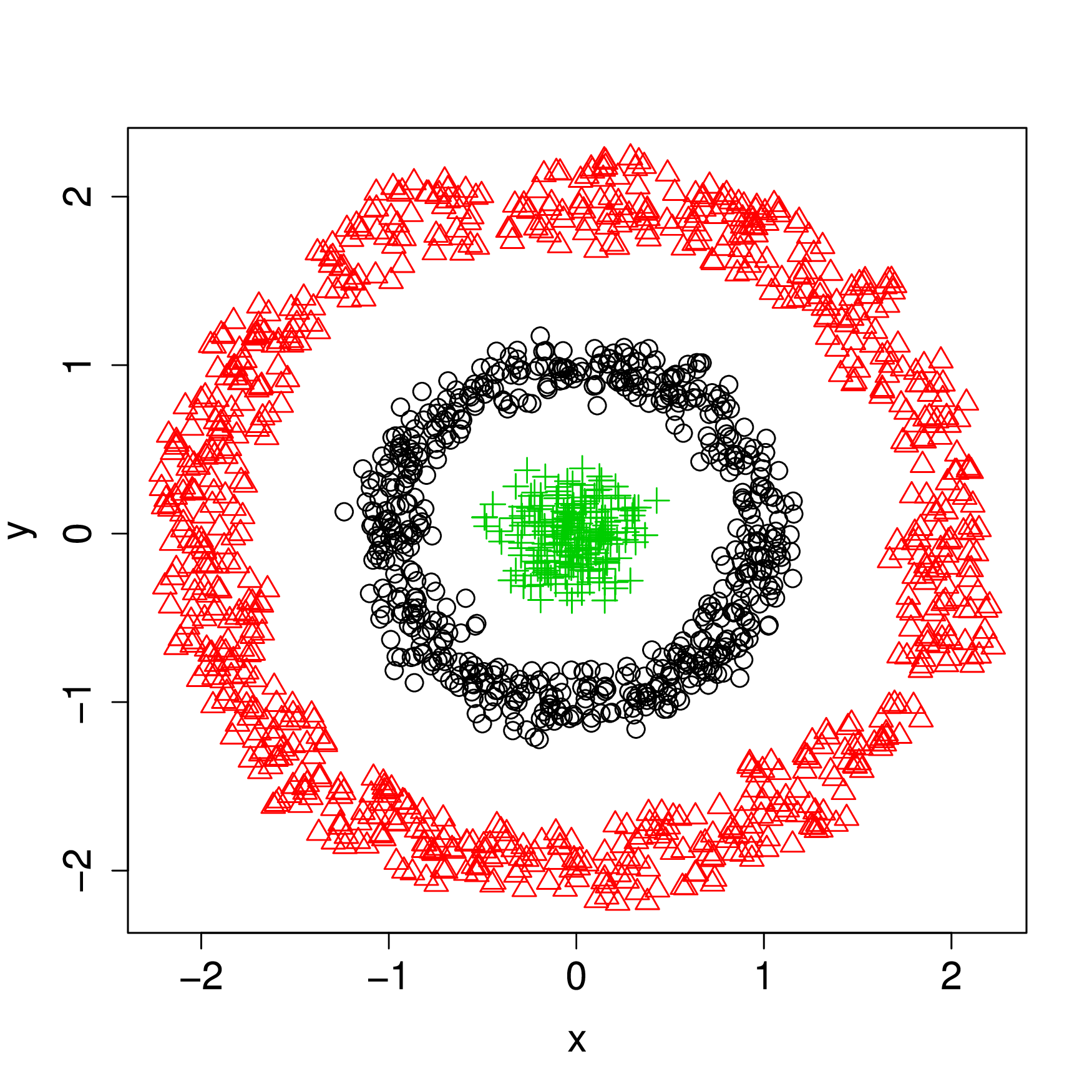} &
\end{tabular}
\end{center}
\caption{\small{\it Top Left}: first 10 eigenvalues, sorted in ascending order.
{\it Top Right}: pairs plots of the first three eigenvectors. It may be seen that the embedded data concentrate around three distinct points in the feature space $\mathbb{R}^3$.
{\it Bottom} Resulting partition obtained by applying a $k$-means algorithm in the feature space. The color scheme is identical to the representation of the eigenvectors (top-right panel). The three groups are accurately recovered.}
\label{fig:clusters}
\end{figure}

\begin{figure}
\begin{center}
\includegraphics[width=0.5\linewidth]{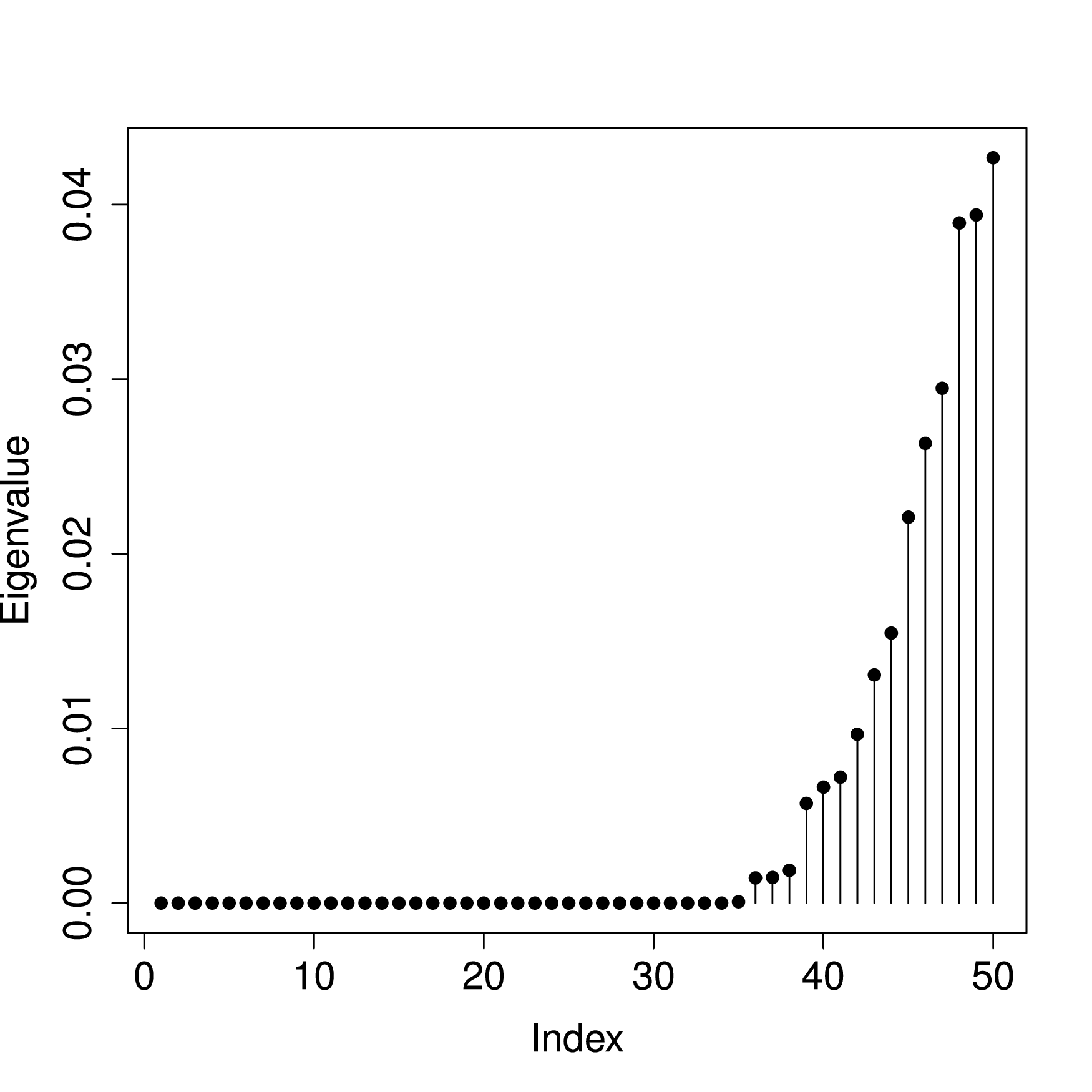}
\end{center}
\caption{\small First 50 eigenvalues of the standard spectral clustering
  algorithm, applied on the initial data set, i.e., without level set
  pre-processing. A total of 35 eigenvalues are found equal to zero,
  which leads to 35 inhomogeneous groups, indicating failure of the
  standard spectral clustering algorithm. }
\label{fig:sceigen}
\end{figure}

We consider a mixture density on $\mathbb{R}^2$ with four components
corresponding to random variables $X_1,\dots,X_4$ where 
\begin{itemize}
\item[(i)] $X_1 \sim \mathcal{N}(0,\sigma_1^2\mathbf{I})$ with $\sigma_1=0.2$ ;
\item[(ii)] $X_2 = R_2 (\cos\theta_2,\sin\theta_2)$ where
  $\theta_2\sim\mathcal{U}([0;2\pi])$ and
  $R_2\sim\mathcal{N}(1,0.1^2)$ ;
\item[(iii)] $X_3 = R_3 (\cos\theta_3,\sin\theta_3)$ where
  $\theta_3\sim\mathcal{U}([0;2\pi])$ and
  $R_3\sim\mathcal{N}(2,0.2^2)$ ;
\item[(iv)] $X_4 \sim \mathcal{U}([-3;3]\times[-3;3])$.
\end{itemize}
The proportions of the components in the mixture are taken as $10\%$,
$32\%$, $53\%$ and $5\%$, respectively.  The fourth component ($X_4$)
represents a uniform background noise.

A random sample of size $n=1,900$ has been simulated according to the
mixture.  Points are displayed in Figure~\ref{fig:points} (left).  A
nonparametric kernel density estimate, with a Gaussian kernel, has
been adjusted to the data.  The bandwidth parameter of the density
estimate has been selected automatically with cross-validation.  A
level $t=0.0444$ has been selected such that $85\%$ of the simulated
points are extracted, i.e., $85\%$ of the observations fall in
$\mathcal{L}_n(t)$.  The extracted and discarded points are displayed
in Figure~\ref{fig:points} (right).
The number of extracted points is equal to $1,615$.
\\

The spectral clustering has been applied to the $1,615$ extracted
points, with the similarity function
\[
k(x)=\exp(-1/(1-\|x\|)^2)\mathbf 1\{\|x\|<1\}.
\]
For numerical stability of the algorithm, we considered the
eigendecomposition of the symmetric matrix $\mathbf{I} -
\mathbf{S}_{n,h}$.  Thus, the eigenspace associated with the
eigenvalue $1$ of the matrix $\mathbf{Q}_{n,h}$ corresponds to the
null space of $\mathbf{I} - \mathbf{S}_{n,h}$.  The scale
parameter $h$ has be empirically chosen equal to $0.25$.  The first 10
eigenvalues of $\mathbf{I} - \mathbf{S}_{n,h}$ are represented in
Figure~\ref{fig:clusters} (top-left).  Three eigenvalues are found
equal to zero, indicating three distinct groups.  The data is then
embedded in $\mathbb{R}^3$ using the three eigenvectors of the null
space of $\mathbf{I}-\mathbf{S}_{n,h}$, and the data is partitioned in
this space using a $k$-means clustering algorithm.  Pair plots of
three eigenvectors of the null space are displayed in
Figure~\ref{fig:clusters}.  It may be observed that the embedded data
are concentrated around three distinct points in the feature space.
Applying a $k$-means algorithm in the feature space leads to the
partition represented in Figure~\ref{fig:clusters}.  Note that
observations considered as background noise are the discarded points
belonging to the complement of $\mathcal{L}_n(t)$.  In this example,
our algorithm is successful at recovering the three expected groups.
\\

As a comparison, we applied the standard spectral clustering algorithm
to the initial data set of size $n=1,900$.  In this case, $35$
eigenvalues are found equal to zero (Figure~\ref{fig:sceigen}).
Applying a $k$-means clustering algorithm in the embedding space
$\mathbb{R}^{35}$ leads to 35 inhomogeneous groups (not displayed
here), none of which corresponds roughly to the expected groups (the
two circular bands and the inner circle).  This failure of the
standard spectral clustering algorithm is explained by the presence of
the background noise which, when unfiltered, perturbs the formation of
distinct groups.  While there remains multiple important questions, in
particular regarding the choice of the parameter $h$, these
simulations illustrate the added value of combining a spectral
clustering algorithm with level-set techniques.

\section{Proof of the convergence of $\widehat{Q}_{n,h}$ 
(Theorem~\ref{theo:converge})}
\label{section:convergence}

\subsection{Preliminaries}
\label{sub:preliminaries}
Let us start with the following simple lemma.
\begin{lem}
\label{lem:an}
Let $\{A_n\}_{n\ge 0}$ be a decreasing sequence of Borel sets in
$\mathbb{R}^d$, with limit $A_\infty=\cap_{n\ge0} A_n$.  If
$\mu(A_\infty)=0$, then
$$\mathbb{P}_n A_n=\frac 1n 
\sum_{i=1}^n \mathbf 1\{X_i\in A_n\}  \to 0 
\quad \text{almost surely as $n\to\infty$,}$$
where $\mathbb{P}_n$ is the empirical measure associated with the
random sample $X_1,\dots,X_n$. 
\end{lem}

\noindent\textbf{Proof.}
First, note that $\lim_n \mu(A_n)=\mu(A_\infty)$.  Next, fix an
integer $k$.  For all $n\geq k$, $A_n\subset A_k$ and so
$\mathbb{P}_n A_n \leq \mathbb{P}_n A_k$.  But $\lim_n
\mathbb{P}_n A_k=\mu(A_k)$ almost surely by the law of large numbers.
Consequently $\limsup_n \mathbb{P}_n A_n\le \mu(A_k)$ almost surely.  Letting
$k\to\infty$ yields 
$$\limsup_n  \mathbb{P}_n A_n \le \mu(A_\infty)=0,$$
which concludes the proof since $\mathbb{P}_n A_n\ge 0$.
\nopagebreak\hfill$\Box$\\

The operator norm convergence that we expect to prove is a uniform law
of large number. The key argument is the fact that the classes of
functions of the following lemma are Glivenko-Cantelli.
Let $g$ be a function defined on some subset $\mathcal{D}$ of
$\mathbb{R}^d$, and let $\mathcal{A}$ be a subset of $\mathcal{D}$.
In what follows, for all $x\in\mathbb{R}^d$, the notation
$g(x)\mathbf{1}_\mathcal{A}(x)$ stands for $g(x)$ if
$x\in\mathcal{A}$ and $0$ otherwise.
\begin{lem}
\label{lem:gc}
1. The two collections of functions
\begin{align*}
  \mathcal{F}_1 &:= \big\{y\mapsto
  k_h(y-x)\mathbf{1}_{\mathcal{L}(t)}(y)\,
  :\,x\in\mathcal{L}(t-\varepsilon_0)\big\},
  \\
  \mathcal{F}_2 &:=\big\{y\mapsto
  D_x k_h(y-x)\mathbf{1}_{\mathcal{L}(t)}(y)\,
  :\,x\in\mathcal{L}(t-\varepsilon_0)\big\},
\end{align*}
are Glivenko-Cantelli, where $D_x k_h$ denotes the differential of
$k_h$.

2. Let $r:\mathcal{L}(t)\times\mathbb{R}^d$ be a continuously
differentiable function such that
\\
\textit{(i)} there exists a compact $\mathcal{K}\subset\mathbb{R}^d$
such that $r(x,y)=0$ for all $(x,y)\in\mathcal{L}(t)\times K^c$;
\\
\textit{(ii)} $r$ is uniformly bounded on $\mathcal L(t)\times
\mathbb{R}^d$, i.e. $\|r\|_\infty < \infty$.
\\
Then the collection of functions
$$\mathcal{F}_3:=
\Big\{ y\mapsto r(x,y)g(y)\mathbf{1}_{\mathcal{L}(t)}(y)
\,:\,x\in\mathcal{L}(t),\, \|g\|_{W(\mathcal{L}(t))}\leq 1\Big\}$$
is Glivenko-Cantelli.
\end{lem}
\noindent\textbf{Proof.}
1.  Clearly $\mathcal{F}_1$ has an integrable
envelope since $k_h$ is uniformly bounded.  Moreover, for each fixed
$y$, the map $x\mapsto k_h(y-x)\mathbf{1}_{\mathcal{L}(t)}(y)$ is
continuous, and $\mathcal{L}(t-\varepsilon_0)$ is compact.  Hence for
each $\delta>0$, using a finite covering of
$\mathcal{L}(t-\varepsilon_0)$, it is easy to construct finitely many
$L_1$ brackets of size at most $\delta$ whose union cover
$\mathcal{F}_1$; see e.g., Example 19.8 in \cite{vandervaart1}. So
$\mathcal{F}_1$ is Glivenko-Cantelli.  Since $k_h$ is
continuously differentiable and with compact support, the same
arguments apply to each component of $D_x k_h$, and so $\mathcal{F}_2$
is also a Glivenko-Cantelli class.
\\

2.  Set $\mathcal{R} = \{y\mapsto
r(x,y)\,:\,x\in\mathcal{L}(t)\}$. 
First, since $r$ is continuous on the
compact set $\mathcal{L}(t)\times\mathcal{K}$, it is uniformly continuous.
So a finite covering of $\mathcal{R}$ of arbitrary size in the
supremum norm may be obtained from a finite covering of
$\mathcal{L}(t)\times\mathcal{K}$.  Hence $\mathcal{R}$ has finite
entropy in the supremum norm.
Second, set $\mathcal{G} = \{y\mapsto
g(y)\mathbf{1}_{\mathcal{L}(t)}(y)\,:\, \|g\|_{W(\mathcal{L}(t))}\leq
1\}$.  Denote by $\mathcal{X}$ the convex hull of $\mathcal{L}(t)$,
and consider the collection of functions $\tilde{\mathcal{G}} =
\{\tilde{g}:\mathcal{X}\to\mathbb{R}\,:\,\|\tilde{g}\|_{W(\mathcal{X})}\leq
1\}$.  Then $\tilde{\mathcal{G}}$ has finite entropy in the supremum
norm; see \cite{kolmo} and \cite{vandervaart2}.  Using the surjection
$\tilde{\mathcal{G}}\to\mathcal{G}$ carrying $\tilde{g}$ to
$\big(\tilde{g}\mathbf{1}_{\mathcal{L}(t)}\big)$, that $\mathcal{G}$
has finite entropy in the supremum norm readily follows.  To conclude
the proof, since both $\mathcal{R}$ and $\mathcal{G}$ are uniformly
bounded, a finite covering of $\mathcal{F}_3$ of arbitrary size
$\delta$ in the supremum norm may be obtained from finite coverings of
$\mathcal{R}$ and $\mathcal{G}$, which yields a finite covering of
$\mathcal{F}_3$ by $L_1$ brackets of size at most $2\delta$.  So
$\mathcal{F}_3$ is a Glivenko-Cantelli class.
\nopagebreak\hfill$\Box$
\\

We recall that the limit operator $Q_h$ is given by
\eqref{eq:Qh}.
The following lemma gives useful bounds on $K_h$ and $q_h$, both defined in
\eqref{eq:qh}.
\begin{lem}
\label{lem:qhqnh}
1. The function $K_h$ is uniformly bounded from below by some positive
number on $\mathcal{L}(t-\varepsilon_0) $, i.e., $\inf\{ K_h(x):\,
x\in\mathcal{L}(t-\varepsilon_0)\}>0$;
\\
2. The kernel $q_h$ is uniformly bounded, i.e., $\|q_h\|_\infty
<\infty$;
\\
3. The differential of $q_h$ with respect to $x$ is uniformly bounded
on $\mathcal{L}(t-\varepsilon_0)\times\mathbb{R}^d $, i.e.,
$\sup\big\{ \|D_xq_h(x,y)\|:(x,y)\in
\mathcal{L}(t-\varepsilon_0)\times\mathbb{R}^d\big\}<\infty$;
\\
4. The Hessian of $q_h$ with respect to $x$ is uniformly
bounded on $\mathcal{L}(t-\varepsilon_0)\times\mathbb{R}^d $, i.e.,
$\sup \big\{\|D_x^2q_h(x,y)\|:(x,y)\in
\mathcal{L}(t-\varepsilon_0)\times\mathbb{R}^d \big\}<\infty$.
\end{lem}
\noindent{\bf Proof.}
First observe that the statements 2, 3 and 4
are immediate consequences of statement 1 together
with the fact that the function $k_h$ is of class $\mathcal{C}^2$ with
compact support, which implies that $k_h(y-x)$, $D_xk_h(y-x)$, and
$D_x^2k_h(y-x)$ are uniformly bounded.

To prove statement 1, note that $K_h$ is continuous and that $K_h(x)>0$
for all $x\in\mathcal{L}(t)$. Set
\[
\alpha(\varepsilon_0)=
\inf\big\{\|D_xf(x)\|;\,
x\in\mathcal{L}_{t-\varepsilon_0}^t\big\}.
\]
Let
$(x,y)\in\mathcal{L}_{t-\varepsilon_0}^t\times\partial\mathcal{L}(t)$. Then
\[
\varepsilon_0 \ge f(y)-f(x) \ge \alpha(\varepsilon_0)\|y-x\|.
\]
Thus,
$\|y-x\|\leq 
{\varepsilon_0}/{\alpha(\varepsilon_0)}$
and so
\[
\mathrm{dist}\big(x,\mathcal{L}(t)\big)
\leq 
\frac{\varepsilon_0}{\alpha(\varepsilon_0)}, \quad
\text{for all }x\in\mathcal L_{t-\varepsilon_0}^t.
\]
Recall from \eqref{eq:vareps0}  that
$
{h}/{2} > {\varepsilon_0}/{\alpha(\varepsilon_0)}$.
Consequently, for all $x\in\mathcal L(t-\varepsilon_0)$, the set
$(x+hB/2)\cap \mathcal L(t)$ contains a non-empty, open set
$U(x)$.
Moreover $k_h$ is bounded from below by some positive number on
$hB/2$ by Assumption 2.  Hence $K_h(x)>0$ for all $x$ in
$\mathcal{L}(t-\varepsilon_0)$ and point 1 follows from the continuity
of $K_h$ and the compactness of $\mathcal{L}(t-\varepsilon_0)$.
\nopagebreak\hfill$\Box$
\\

In order to prove the convergence of $\widehat{Q}_{n,h}$ to $Q_h$, we
also need to study the uniform convergence of $K_{n,h}$, given in
\eqref{eq:qnh}. Lemma~\ref{lem:knh} controls the difference
between $K_{n,h}$ and $K_h$, while Lemma~\ref{lem:ratios} controls the
ratio of $K_h$ over $K_{n,h}$.
\begin{lem}
\label{lem:knh}
As $n\to\infty$, almost surely,
\\
1.
$\displaystyle\sup_{x\in\mathcal{L}(t-\varepsilon_0)}\Big|K_{n,h}(x) -
K_h(x)\Big| \to 0$ ~ and
\\
2.
$\displaystyle\sup_{x\in\mathcal{L}(t-\varepsilon_0)}\Big|D_xK_{n,h}(x)
- D_xK_h(x)\Big| \to 0.$ 
\end{lem}
\noindent\textbf{Proof.}
Let 
\begin{align*}
K_{n,h}^\dag(x)&:=\frac{1}{n\mu(\mathcal L(t))}
\sum_{i=1}^n k_h(X_i-x)\mathbf 1_{\mathcal L_n(t)}(X_i),
\\
K_{n,h}^{\dag\dag}(x)&:=\frac{1}{n\mu(\mathcal L(t))}
\sum_{i=1}^n k_h(X_i-x)\mathbf 1_{\mathcal L(t)}(X_i).
\end{align*}
Let us start with the inequality
\begin{equation}
\label{eq:knh4}
\Big|K_{n,h}(x) - K_h(x)\Big|\le 
\Big|K_{n,h}(x) - K_{n,h}^\dag(x)\Big| +
\Big|K_{n,h}^\dag(x) - K_h(x)\Big|,
\end{equation}
for all $x\in\mathcal{L}(t-\varepsilon_0)$.
Using the inequality
\[
\Big|K_{n,h}(x) - K_{n,h}^\dag(x)\Big| \le \left| \frac{n}{j(n)} -
  \frac{1}{\mu(\mathcal L(t))} \right| \, \|k_h\|_\infty
\]
we conclude that the first term in (\ref{eq:knh4}) tends to 0
uniformly in $x$ over $\mathcal{L}(t-\varepsilon_0)$ with probability
one as $n\to\infty$, since $j(n)/n \to \mu\big(\mathcal{L}(t)\big)$
almost surely, and since $k_h$ is bounded on $\mathbb{R}^d$.\\

Next, for all $x\in\mathcal{L}(t-\varepsilon_0)$, we have
\begin{equation}\label{eq:knh5}
\Big|K_{n,h}^\dag(x) - K_h(x)\Big| \le 
\Big|K_{n,h}^\dag(x) - K_{n,h}^{\dag\dag}(x)\Big| +
\Big|K_{n,h}^{\dag\dag}(x) - K_h(x)\Big|.
\end{equation}
The first term in \eqref{eq:knh5} is bounded by
\begin{align*}
\Big|K_{n,h}^\dag(x) - K_{n,h}^{\dag\dag}(x)\Big| 
&   \le  \frac{\|k_h\|_\infty}{\mu\big(\mathcal{L}(t)\big)}\frac{1}{n} 
  \left|
    \sum_{i=1}^n \Big\{\mathbf{1}_{\mathcal{L}_n(t)}(X_i) -
    \mathbf{1}_{\mathcal{L}(t)}(X_i)\Big\}
  \right|\\
& =  \frac{\|k_h\|_\infty}{\mu\big(\mathcal{L}(t)\big)}\frac{1}{n} 
\sum_{i=1}^n \mathbf{1}_{\mathcal{L}_n(t)\Delta\mathcal{L}(t)}(X_i),
\end{align*}
where $\mathcal{L}_n(t)\Delta\mathcal{L}(t)$ denotes the symmetric
difference between $\mathcal{L}_n(t)$ and $\mathcal{L}(t)$.  Recall
that, on the event $\Omega_n$, $\mathcal L(t-\epsilon_n) \subset
\mathcal L_n(t) \subset \mathcal{L} (t-\varepsilon_n)$.  Therefore
$\mathcal{L}_n(t)\Delta\mathcal{L}(t) \subset
\mathcal{L}_{t-\varepsilon_n}^{t+\varepsilon_n}$ on $\Omega_n$, and so
\begin{equation*}
 0\le \frac 1n \left| 
   \sum_{i=1}^n \Big\{\mathbf{1}_{\mathcal{L}_n(t)}(X_i) -
    \mathbf{1}_{\mathcal{L}(t)}(X_i)\Big\}
 \right|\mathbf 1_{\Omega_n}
   \le\frac 1n \sum_{i=1}^n\mathbf{1}_{A_n}(X_i),
\end{equation*}
where $A_n = \mathcal{L}_{t-\varepsilon_n}^{t+\varepsilon_n}$.  Hence
by Lemma~\ref{lem:an}, and since $\mathbf{1}_{\Omega_n}\to1$ almost
surely as $n\to\infty$, the first term in (\ref{eq:knh5}) converges to
$0$ with probability one as $n\to\infty$.\\

Next, since the collection $\big\{y\mapsto k_h(y-x)
\mathbf{1}_{\mathcal{L}(t)}(y)\,:\,x\in\mathcal{L}(t-\varepsilon_0)\big\}$
is Glivenko-Cantelli by Lemma~\ref{lem:gc}, we conclude that
$$\sup_{x\in \mathcal L(t-\varepsilon_0)}
\Big|K_{n,h}^{\dag\dag}(x) - K_h(x)\Big| \to 0,$$ with probability one as
$n\to\infty$. This concludes the proof of the first statement.\\

The second statement may be proved by developing similar arguments,
with $k_h$ replaced by $D_xk_h$, and by noting that the collection of
functions $\big\{y\mapsto D_x k_h(y-x)
\mathbf{1}_{\mathcal{L}(t)}(y)\,:\,x\in\mathcal{L}(t-\varepsilon_0)\big\}$
is also Glivenko-Cantelli by Lemma~\ref{lem:gc}.
\nopagebreak\hfill$\Box$
\\

\begin{lem}
\label{lem:ratios}
As $n\to\infty$, almost surely,\\
1. $\displaystyle \sup_{x\in\mathcal{L}(t)} 
\bigg|\frac{K_h\big(\varphi_n(x)\big)}{K_{n,h}\big(\varphi_n(x)\big)} -1 \bigg|
\to 0,$ and\\

2. $\displaystyle \sup_{x\in\mathcal{L}(t)} \bigg\| 
D_x\bigg[
\frac{K_h\big(\varphi_n(x)\big)}{K_{n,h}\big(\varphi_n(x)\big)} 
\bigg]  \bigg\|  \to 0.$
\end{lem}

\noindent{\bf Proof.}
First of all, $K_h$ is uniformly continuous on $\mathcal
L(t-\varepsilon_0)$ since $K_h$ is continuous and since $\mathcal
L(t-\varepsilon_0)$ is compact.  Moreover, $\varphi_n$ converges
uniformly to the identity map of $\mathcal L(t)$ by
Lemma~\ref{lem:levelsets}.  Hence
$$\sup_{x\in\mathcal{L}(t)} \big|K_h\big(\varphi_n(x)\big) - K_h(x)\big|
\to 0 \quad\text{as $n\to\infty$,}$$
and since $K_{n,h}$ converges uniformly to $K_h$ with probability one
as $n\to\infty$ by Lemma~\ref{lem:knh}, this proves 1.\\

We have
\begin{multline*}
D_x\left[ \frac{K_h\big(\varphi_n(x)\big)}{K_{n,h}\big(\varphi_n(x)\big)}
\right] 
 =
\Big[K_{n,h}\big(\varphi_n(x)\big)\Big]^{-2} D_x\varphi_n(x)\\
 \times
\Big[
 K_{n,h}\big(\varphi_n(x)\big) D_xK_h\big(\varphi_n(x)\big) 
- K_h\big(\varphi_n(x)\big)D_xK_{n,h}\big(\varphi_n(x)\big)
\Big].
\end{multline*}
Since $D_x\varphi_n(x)$ converges to the identity matrix $I_d$
uniformly over $x\in\mathcal L(t)$ by Lemma~\ref{lem:levelsets},
$\|D_x\varphi_n(x)\|$ is bounded uniformly over $n$ and $x\in\mathcal
L(t)$ by some positive constant $C_\varphi$.  Furthermore the map
$x\mapsto K_{n,h}(x)$ is bounded from below over $\mathcal L(t)$ by
some positive constant $k_{min}$ independent of $x$ because i)
$\inf_{x\in\mathcal{L}(t-\varepsilon_0)} K_h(x) >0$ by
Lemma~\ref{lem:qhqnh}, and ii)
$\sup_{x\in\mathcal{L}(t-\varepsilon_0)}\big|K_{n,h}(x)-K_h(x)\big|\to
0$ by Lemma~\ref{lem:knh}.  Hence
\begin{equation*}
\left| D_x\bigg[ \frac{K_h\big(\varphi_n(x)\big)}{K_{n,h}\big(\varphi_n(x)\big)} 
\bigg] \right|
\le \frac{C_\varphi}{k_{min}^2}
\Big|
 K_{n,h}(y) D_xK_h(y) 
- K_h(y)D_xK_{n,h}(y)
\Big|,
\end{equation*}
where we have set $y=\varphi_n(x)$ which belongs to $\mathcal L(t-\varepsilon_n)\subset\mathcal L(t-\varepsilon_0)$.
At last, Lemma~\ref{lem:knh} gives
\[
\sup_{y\in\mathcal L(t-\varepsilon_0)}
\Big|
 K_{n,h}(y) D_xK_h(y) 
- K_h(y)D_xK_{n,h}(y)
\Big|\to 0
\quad \text{almost surely},
\]
as $n\to\infty$ which proves 2.
\nopagebreak\hfill$\Box$
\\


We are now almost ready to prove the uniform convergence of empirical
operators. The following lemma is a consequence of
Lemma~\ref{lem:gc}. 
\begin{lem}
\label{lem:rconv}
Let $r:\mathcal{L}(t-\varepsilon_0)\times\mathbb{R}^d\to\mathbb{R}$ be
a continuously differentiable function with compact support such that
\textit{(i)} $r$ is uniformly bounded on $\mathcal L(t-\varepsilon_0)
\times\mathbb R^d $, i.e., $\|r\|_\infty < \infty $, and
\textit{(ii)} the differential $D_xr$ with respect to $x$
is uniformly bounded on $\mathcal L(t-\varepsilon_0)
\times\mathbb R^d$, i.e.,
$\displaystyle
\|D_xr\|_\infty:=\sup\left\{ \|D_xr(x,y)\| :\,
(x,y)\in \mathcal{L}(t-\varepsilon_0)\times\mathbb{R}^d
\right\}
<\infty$.
\\
Define the linear operators $R_n$ and $R$ on $W\big(\mathcal L(t)\big)$
respectively by
\begin{align*}
  R_n g(x) & =\int_{\mathcal{L}_n(t)} r\big(\varphi_n(x),y\big) 
  g\big(\varphi_n^{-1}(y)\big) \mathbb{P}_n^t(dy),
  \\
  R g(x) & =\int_{\mathcal{L}(t)} r(x,y) g(y) \mu^t(dy).
\end{align*}
Then, as $n\to\infty$,
\[
\sup\Big\{ \big\|R_ng -Rg\big\|_\infty\, :\, \|g\|_W\le 1 \Big\}\to 0
\quad \text{almost surely}.
\]
\end{lem}

\noindent{\bf Proof.}
Set
\begin{align*}
  S_ng(x) & :=\frac{1}{\mu(\mathcal L(t))} \frac{1}{n}\sum_{i=1}^n
  r\big(\varphi_n(x),X_i\big) \,
  g\big(\varphi_n^{-1}(X_i)\big) \mathbf 1_{\mathcal L_n(t)}
  (X_i),
  \\
  T_ng(x) & := \frac{1}{\mu\big(\mathcal{L}(t)\big)}\frac 1n 
  \sum_{i=1}^n r\big(\varphi_n(x),X_i\big) g(X_i)
  \mathbf 1_{\mathcal L(t)}(X_i),
  \\
  U_ng(x) & :=  \frac{1}{\mu\big(\mathcal{L}(t)\big)}\frac 1n 
  \sum_{i=1}^n r\big(x,X_i\big) g(X_i)
  \mathbf 1_{\mathcal L(t)}(X_i).
\end{align*}
and consider the inequality
\begin{align}
\big| R_ng(x) - Rg(x)\big| & \le \big| R_ng(x) - S_ng(x)\big| + \big| S_ng(x) - T_ng(x)\big|\notag\\
& \quad +\big|T_ng(x) - U_ng(x)\big| + \big|U_ng(x)-Rg(x)\big|,\label{eq:rconv00}
\end{align}
for all $x\in\mathcal{L}(t)$ and all $g\in W\big(\mathcal{L}(t)\big)$.

The first term in \eqref{eq:rconv00} is bounded uniformly by
\[
\big| R_ng(x) - S_ng(x)\big| 
\leq \bigg|\frac{n}{j(n)} - 
\frac{1}{\mu\big(\mathcal{L}(t)\big)}\bigg| \|r\|_\infty \|g\|_\infty
\]
and since $j(n)/n$ tends to $\mu(\mathcal L(t))$ almost surely as $n\to\infty$, we conclude that
\begin{equation}
\label{eq:rconv01}
\sup\Big\{ \big\|R_ng - S_ng\big\|_\infty \,:\, \|g\|_W \leq 1\Big\} \to 0\quad  \text{a.s. as $n\to\infty$}.
\end{equation}

For the second term in \eqref{eq:rconv00}, we have
\begin{align}
|S_ng(x) - T_ng(x)|  &\le
\frac{\|r\|_\infty}{\mu\big(\mathcal{L}(t)\big)} \frac{1}{n} \sum_{i=1}^n 
\big| g\big(\varphi_n^{-1}(X_i)\big) \mathbf 1_{\mathcal L_n(t)}(X_i) 
- g(X_i)\mathbf 1_{\mathcal L(t)}(X_i) \big|
\notag\\
& = \frac{\|r\|_\infty}{\mu\big(\mathcal{L}(t)\big)} \frac{1}{n} \sum_{i=1}^n g_n(X_i), \label{eq:rconv02}
\end{align}
where $g_n$ is the function defined on the whole space $\mathbb R^d$
by
$$g_n(x) = \Big| g\big(\varphi_n^{-1}(x)\big)\mathbf{1}_{\mathcal{L}_n(t)}(x)- g(x)\mathbf 1_{\mathcal L(t)}(x)\Big|.$$
Consider the partition of $\mathbb R^d$ given by
$
\mathbb R^d = B_{1,n} \cup B_{2,n} \cup B_{3,n} \cup B_{4,n},
$
where 
\[
\begin{array}{ll}
B_{1,n}:=\mathcal L_n(t)\cap \mathcal L(t),
& B_{2,n}:=\mathcal L_n(t)\cap \mathcal L(t)^c,\\
B_{3,n}:=\mathcal L_n(t)^c\cap \mathcal L(t), &
B_{4,n}:=\mathcal L_n(t)^c\cap \mathcal L(t)^c.
\end{array}
\]
The sum over $i$ in \eqref{eq:rconv02} may be split into four parts as
\begin{equation}
\label{eq:rconv03}
\frac 1n \sum_{i=1}^n g_n(X_i) = 
I_1(x,g) +I_2(x,g) +I_3(x,g) +I_4(x,g) 
\end{equation}
where
\[
I_k(x,g):=\frac 1n
\sum_{i=1}^n g_n(X_i) \mathbf 1\{X_i\in B_{k,n}\}.
\]
First, $I_{4,n}(x,g)=0$ since $g_n$ is identically 0 on $B_{4,n}$.
Second,
\begin{equation}
\label{eq:rconv04}
I_2(x,g)+I_3(x,g)\le \|g\|_\infty \frac 1n \sum_{i=1}^n \mathbf
1_{\mathcal L(t)\Delta \mathcal L_n(t)}(X_i)
\end{equation}
Applying Lemma~\ref{lem:an} together with the almost sure convergence of $\mathbf
1_{\Omega_n}$ to 1, we obtain that 
\begin{equation}
\label{eq:rconv05}
\frac 1n \sum_{j=1}^n \mathbf 1_{\mathcal L(t)\Delta \mathcal
  L_n(t)}(X_j)\to 0 
\quad\text{almost surely}.
\end{equation}
Third,
\begin{align}
I_1(x,g) &\le \sup_{x\in \mathcal L(t)}
\bigg|  g\big(\varphi_n^{-1}(x)\big) - g(x)\bigg|\notag\\
&\le \|D_xg\|_\infty \sup_{x\in \mathcal L(t)}\|\varphi_n^{-1}(x)-x\|
\notag\\
& \le \|D_xg\|_\infty \sup_{x\in \mathcal L(t)}\|x-\varphi_n(x)\|\notag\\
& \to 0 \label{eq:rconv06}
\end{align}
as $n\to\infty$ by Lemma~\ref{lem:levelsets}.
Thus, combining
\eqref{eq:rconv02}, \eqref{eq:rconv03}, \eqref{eq:rconv04}, \eqref{eq:rconv05}
and \eqref{eq:rconv06} leads to
\begin{equation}
\label{eq:rconv07}
\sup\Big\{\big\|S_ng - T_n g\big\|_\infty \,: \, \|g\|_W\le 1\Big\}\to 0 \quad \text{a.s. as $n\to\infty$}.
\end{equation}

For the third term in (\ref{eq:rconv00}), using the inequality
\[
\left|r\big(\varphi_n(x),X_i\big)-r\big(x,X_i\big)\right| \le
\|D_xr\|_\infty \sup_{x\in\mathcal L(t)}\|\varphi_n(x)-x\|
\]
we deduce that
\[
\big|T_n g(x) - U_n g(x)\big| \le
\frac{1}{\mu\big(\mathcal{L}(t)\big)} \|g\|_\infty \|D_xr\|_\infty \sup_{x\in\mathcal L(t)}\|\varphi_n(x)-x\|.
\]
and so
\begin{equation}
\label{eq:rconv08}
\sup\Big\{\big\|T_ng - U_n g\big\|_\infty \,: \, \|g\|_W\le 1\Big\}\to 0 \quad \text{a.s. as $n\to\infty$},
\end{equation}
by Lemma~\ref{lem:levelsets}.\\

At last, for the fourth term in (\ref{eq:rconv00}), since the function
$r$ satisfies the conditions of the second statement in
Lemma~\ref{lem:gc}, we conclude by Lemma~\ref{lem:gc} that 
\begin{equation}
\label{eq:rconv09}
\sup\Big\{\big\|U_ng - R g\big\|_\infty \,: \, \|g\|_W\le 1\Big\}\to 0 
\quad \text{a.s. as $n\to\infty$}.
\end{equation}
Finally, reporting (\ref{eq:rconv01}), (\ref{eq:rconv07}) and
(\ref{eq:rconv08}) in (\ref{eq:rconv00}) yields the desired result.
\nopagebreak\hfill$\Box$

\subsection{Proof of Theorem~\ref{theo:converge}}
\label{sub:convergence}
We will prove that, as $n\to\infty$, almost surely,
\begin{equation}
\label{eq:conv01}
\sup \bigg\{\Big\|\widehat{Q}_{n,h}g - Q_hg \Big\|_\infty \,:\,
\|g\|_W\leq 1\bigg\}\to 0
\end{equation}
and
\begin{equation}
\label{eq:conv02}
\sup \bigg\{ \Big\|D_x\big[\widehat{Q}_{n,h}g\big] - D_x\big[Q_hg\big]
\Big\|_\infty \,:\, \|g\|_W\leq 1\bigg\}\to 0
\end{equation}

To this aim, we introduce the operator $\widetilde{Q}_{n,h}$ acting on $W(\mathcal L(t))$ as
\begin{equation*}
\widetilde{Q}_{n,h}g(x) = 
\int_{\mathcal L_n(t)} q_h(\varphi_n(x),y)g\big(\varphi_n^{-1}(y)\big)
\mathbb{P}_n^t(dy).
\end{equation*}

\paragraph{Proof of (\ref{eq:conv01})}

For all $g\in W\big(\mathcal L(t)\big)$, we have
\begin{equation}
\label{eq:conv03}
\big\|\widehat{Q}_{n,h}g - Q_h g \big\|_\infty
\le
\big\|\widehat{Q}_{n,h}g - \widetilde{Q}_{n,h} g \big\|_\infty
+
\big\|\widetilde{Q}_{n,h}g - Q_h g \big\|_\infty.
\end{equation}
First, by Lemma~\ref{lem:qhqnh}, the function $r=q_h$ satisfies the condition in Lemma~\ref{lem:rconv}, so that
\begin{equation}
\label{eq:conv04}
\sup\left\{\|\widetilde{Q}_{n,h}g - Q_h g \|_\infty
:\, \|g\|_W\le 1
\right\}\to 0
\end{equation}
with probability one as $n\to\infty$.\\

Next, since $\|q_h\|_\infty < \infty$ by Lemma~\ref{lem:qhqnh}, there exists a finite
constant $C_h$ such that, 
\begin{equation}
\label{eq:bb1}
\|\widetilde{Q}_{n,h}g\|_\infty\le C_h
\quad\text{for all }n\text{ and all }g\text{ with }
\|g\|_W\le1.
\end{equation}
By definition of $q_{n,h}$, for all $x,y$ in the level set $\mathcal L(t)$, we have
\begin{equation}
\label{eq:conv05}
q_{n,h}(x,y)=\frac{K_h(x)}{K_{n,h}(x)} q_h(x,y).
\end{equation}
So
\begin{align*}
\left| 
\widehat{Q}_{n,h}g(x) - \widetilde{Q}_{n,h} g(x)
\right|
& = \left|\frac{K_n\big(\varphi_n(x)\big)}{K_{n,h}\big(\varphi_n(x)\big)}-1 \right|\,
\left|\widetilde{Q}_{n,h} g(x) \right|
\\
& \le C_h 
\sup_{x\in \mathcal L(t)}
\left|\frac{K_n\big(\varphi_n(x)\big)}{K_{n,h}\big(\varphi_n(x)\big)}-1 \right|,
\end{align*}
where $C_h$ is as in \eqref{eq:bb1}.
Applying Lemma~\ref{lem:ratios} yields
\begin{equation}
\label{eq:conv06}
\sup\left\{\| \widehat{Q}_{n,h}g - \widetilde{Q}_{n,h}g \|_\infty
:\, \|g\|_W\le 1
\right\}\to 0
\end{equation}
with probability one as $n\to\infty$.
Reporting (\ref{eq:conv04}) and (\ref{eq:conv06}) in (\ref{eq:conv03}) proves (\ref{eq:conv01}).

\paragraph{Proof of (\ref{eq:conv02})}

We have
\begin{multline}
\label{eq:der01}
\bigg\|D_x\Big[\widehat{Q}_{n,h}g \Big] -
D_x\Big[{Q}_{h}g \Big]\bigg\|_\infty
\\
\le
\bigg\|D_x\Big[\widehat{Q}_{n,h}g \Big] -D_x\Big[\widetilde{Q}_{h}g
  \Big]\bigg\|_\infty
+
\bigg\|D_x\Big[\widetilde{Q}_{n,h}g \Big] -D_x\Big[{Q}_{h}g
  \Big]\bigg\|_\infty.
\end{multline}
The second term in (\ref{eq:der01}) is bounded by
\begin{equation*}
\bigg\|D_x\Big[\widetilde{Q}_{n,h}g \Big] -D_x\Big[{Q}_{h}g
  \Big]\bigg\|_\infty \le
\big\|D_x\varphi_n\big\|_\infty \, \big\|R_ng-Rg \big\|_\infty,
\end{equation*}
where
\begin{align*}
R_ng(x) & :=
\int_{\mathcal L_n(t)} (D_xq_h)(\varphi_n(x),y) g\big(\varphi_n^{-1}(y)\big)
\mathbb{P}_n^t(dy)\quad\text{and}
\\
Rg(x) & :=
\int_{\mathcal L(t)} (D_xq_h)(\varphi_n(x),y) g\big(\varphi_n^{-1}(y)\big)
\mu^t(dy).
\end{align*}
By lemma~\ref{lem:levelsets}, $x\mapsto D_x\varphi_n(x)$ converges
to the identity matrix $I_d$ of $\mathbb R^d$, uniformly in $x$ over
$\mathcal L(t)$.
So $\|D_x\varphi_n(x)\|$ is bounded by some finite
constant $C_\varphi$ uniformly over $n$ and $x\in\mathcal L(t)$ and
\begin{equation*}
\bigg\|D_x\Big[\widetilde{Q}_{n,h}g \Big] -D_x\Big[{Q}_{h}g
  \Big]\bigg\|_\infty \le
C_\varphi \big\|R_ng-Rg \big\|_\infty.
\end{equation*}
By Lemma~\ref{lem:qhqnh}, the map $r:(x,y)\mapsto D_x q_h(x,y)$
satisfies the conditions in Lemma~\ref{lem:rconv}. Thus,
$\|R_ng-Rg \|_\infty$
converges to 0 almost surely, uniformly over $g$ in the unit ball of
$W(\mathcal L(t))$, and we deduce that
\begin{equation}
\label{eq:der02}
\sup\Bigg\{ \bigg\|D_x\Big[\widetilde{Q}_{n,h}g \Big] -D_x\Big[{Q}_{h}g
  \Big]\bigg\|_\infty : \|g\|_W\le 1\Bigg\} 
\to 0 \quad \text{a.s. as $n\to\infty$.}
\end{equation}

For the first term in \eqref{eq:der01}, observe first that there
exists a constant $C'_h$ such that, for all $n$ and all $g$ in the
unit ball of $W\big(\mathcal L(t)\big)$,
\begin{equation}
\label{eq:bb2}
\left\|R_{n,h}g \right\|_\infty \le C'_h,
\quad\text{for all }n\text{ and all }g\text{ with }
\|g\|_W\le1,
\end{equation}
by Lemma~\ref{lem:qhqnh}.\\

On the one hand, we have
\begin{align*}
D_x\big[q_{n,h}(\varphi_n(x),y)\big] & =
\frac{K_h\big(\varphi_n(x)\big)}{K_{n,h}\big(\varphi_n(x)\big)} D_x\varphi_n(x) (D_xq_h)\big(\varphi_n(x),y\big)\\
&\quad +
D_x\left[\frac{K_h\big(\varphi_n(x)\big)}{K_{n,h}\big(\varphi_n(x)\big)}\right]
q_h\big(\varphi_n(x),y\big).
\end{align*}
Hence,
\[
D_x\left[ \widehat{Q}_{n,h}g(x)\right]=
\frac{K_h\big(\varphi_n(x)\big)}{K_{n,h}\big(\varphi_n(x)\big)} D_x\varphi_n(x) R_ng(x)
+
D_x\left[\frac{K_h\big(\varphi_n(x)\big)}{K_{n,h}\big(\varphi_n(x)\big)}\right]
\widetilde{Q}_{n,h}g(x).
\]
\medskip

On the other hand, since
$D_x\big[q_h\big(\varphi_n(x),y\big)\big]=D_x\varphi_n(x) (D_xq_h)\big(\varphi_n(x),y\big)$,
\[
D_x\left[\widetilde{Q}_{n,h}g(x)\right]
= D_x\varphi_n(x) R_ng(x).
\]
Thus,
\begin{multline*}
  D_x\Big[\widehat{Q}_{n,h}g(x) \Big] -D_x\Big[\widetilde{Q}_{h}g(x)
  \Big]= 
D_x\left[\frac{K_h\big(\varphi_n(x)\big)}{K_{n,h}\big(\varphi_n(x)\big)}\right]
\widetilde{Q}_{n,h}g(x)
\\ 
+\left(\frac{K_h\big(\varphi_n(x)\big)}{K_{n,h}\big(\varphi_n(x)\big)} - 1
\right)D_x\varphi_n(x) R_ng(x).
\end{multline*}
Using the inequalities \eqref{eq:bb1} and \eqref{eq:bb2}, we obtain
\begin{align*}
\left\|D_x\Big[\widehat{Q}_{n,h}g \Big] -D_x\Big[\widetilde{Q}_{h}g
  \Big]\right\|_\infty & \le
C_h \sup_{x\in\mathcal L(t)} 
\left|D_x\left[\frac{K_h\big(\varphi_n(x)\big)}{K_{n,h}\big(\varphi_n(x)\big)}\right]\right| \notag\\
& \quad + C'_h C_\varphi
\sup_{x\in\mathcal L(t)}\left|
\frac{K_h\big(\varphi_n(x)\big)}{K_{n,h}\big(\varphi_n(x)\big)} -1
\right|.
\end{align*}
and by applying Lemma~\ref{lem:ratios}, we deduce that
\begin{equation}
\label{eq:der03}
\sup\Bigg\{ \bigg\|D_x\Big[\widehat{Q}_{n,h}g \Big] -D_x\Big[\widetilde{Q}_{h}g
  \Big]\bigg\|_\infty \,:\, \|g\|_W\leq 1\Bigg\} \to 0 \quad \text{a.s. as $n\to\infty$.}
\end{equation}
Reporting \eqref{eq:der02} and \eqref{eq:der03} in \eqref{eq:der01} proves \eqref{eq:conv02}.
\nopagebreak\hfill$\Box$\\


\section{Proof of Corollary~\ref{theo:strong} }
\label{sec:strong}

Let us start with the following proposition, which relates the
spectrum of the functional operator $\widehat{Q}_{n,h}$ with the
one of the matrix $\mathbf{Q}_{n,h}$. 

\begin{pro}
  \label{pro:conjugate}
  On $\Omega_n$, we have
  $
  \pi_n\Phi_n\widehat{Q}_{n,h}=\mathbf{Q}_{n,h}\pi_n\Phi_n
  $
  and the spectrum of the functional operator
  $\widehat{Q}_{n,h} $ is 
  $
  \sigma(\widehat{Q}_{n,h})=\{0\}\cup
  \sigma(\mathbf{Q}_{n,h}).
  $
\end{pro}
\noindent\textbf{Proof.}
Recall that the evaluation map $\pi_n$ defined in \eqref{eq:evalmap}
is such that $\mathbf Q_{n,h}\pi_n=\pi_n Q_{n,h}$, and that, on
$\Omega_n$, $\widehat{Q}_{n,h}=\Phi_nQ_{n,h}\Phi_n^{-1}$. Moreover,
since $\widehat{Q}_{n,h}$ and $Q_{n,h}$ are conjugate, their
spectra are equal. Thus, there remains to show that
$\sigma({Q}_{n,h})=\{0\}\cup \sigma(\mathbf{Q}_{n,h})$.

Remark that $Q_{n,h}$ is a finite rank operator, and that its range is
spanned by the maps $x\mapsto q_{n,h}(x,X_j)$, for $j\in J(n)$. Thus
its spectrum is composed of $0$ and its eigenvalues.  By the relation
$\mathbf{Q}_{n,h}\pi_n=\pi_n Q_{n,h}$, it immediately follows that if
$g$ is an eigenfunction of $Q_{n,h}$ with eigenvalue $\lambda$, then
$V=\pi_n(g)$ is an eigenvector of $\mathbf Q_{n,h}$ with eigenvalue
$\lambda$.  Conversely, if $\{V_j\}_{j}$ is an eigenvector of $\mathbf
Q_{n,h}$, then with some easy algebra, it may be verified that the
function $g$ defined by
\[
g(x):= \sum_{j\in J(n)} V_j \,q_{n,h}(x,X_j)
\]
is an eigenfunction of $Q_{n,h}$ with the same eigenvalue.
\nopagebreak\hfill$\Box$


The spectrum of $Q_h$ may be decomposed as
$\sigma(Q_h)=\sigma_1(Q_h)\cup \sigma_2(Q_h)$, where
$\sigma_1(Q_h)=\{1\}$ and where $\sigma_2(Q_h)=\sigma(Q_h)\setminus
\{1\}$. Since $1$ is an isolated eigenvalue, there exists $\eta_0$ in
the open interval $(0;1)$ such that $\sigma(Q_h)\cap \{z\in\mathbb
C:|z-1|\le\eta_0\}$ is reduced to the singleton $\{1\}$. Moreover, $1$
is an eigenvalue of $Q_h$ of multiplicity $\ell$, by
proposition~\ref{pro:harmonic}. Hence by
Theorem~\ref{theo:separation}, $W\big(\mathcal{L}(t)\big)$ decomposes
into $W\big(\mathcal{L}(t)\big)=M_1\oplus M_2$ where $\dim(M_1)=\ell$.\\

Split the spectrum of $\widehat{Q}_{n,h}$ as
$\sigma\big(\widehat{Q}_{n,h}\big)=
\sigma_1\big(\widehat{Q}_{n,h}\big)\cup\sigma_2\big(\widehat{Q}_{n,h}\big)$,
where 
\[
\sigma_1\big(\widehat{Q}_{n,h}\big) =
\sigma\big(\widehat{Q}_{n,h}\big)\cap \big\{z\in\mathbb C:|z-1|<
\eta_0\big\}.
\]
By Theorem~\ref{theo:separation}, this decomposition of the spectrum
of $\widehat{Q}_{n,h}$ yields a decomposition of $W\big(\mathcal
L(t)\big)$ as $W\big(\mathcal L(t)\big)=M_{n,1}\oplus M_{n,2}$, where
$M_{n,1}$ and $M_{n,2}$ are stable subspaces under
$\widehat{Q}_{n,h}$. 
Statements 4 and
6 of Theorem~\ref{theo:continuous}, together with
Proposition~\ref{pro:conjugate}, gives the following convergences.

\begin{pro}
\label{pro:eigen}
The first $\ell$ eigenvalues $\lambda_{n,1}, \lambda_{n,2},\ldots,
\lambda_{n,\ell}$ of $\mathbf Q_{n,h}$ converge to 1 almost surely as
$n\to\infty$ and there exists $\eta_0>0$ such that, for all $j>\ell$,
$\lambda_{n,j}$ belongs to $\{z:|z-1|\ge \eta_0\}$ for $n$ large
enough, with probability one.
\end{pro}

In addition to the convergence of the eigenvalues of
$\mathbf{Q}_{n,h}$, the convergence of eigenspaces also holds.  More
precisely, let $\Pi$ be the projector on $M_1$ along $M_2$ and $\Pi_n$
the projector on $M_{n,1}$ along $M_{n,2}$. Statements 2, 3, 5 and 6
of Theorem~\ref{theo:continuous} leads to

\begin{pro}
\label{pro:eigen}
 $\Pi_n$ converges to $\Pi$ in operator norm almost surely and
the dimension of $M_{n,1}$ is $\ell$ for all large enough $n$.
\end{pro}

Denote by $E_{n,1}$ the subspace of $\mathbb R^{J(n)}$ spanned by
the eigenvectors of $\mathbf{Q}_{n,h}$ corresponding to the
eigenvalues $\lambda_{n,1}$, \ldots $\lambda_{n,\ell}$. If $n$ is
large enough, we have the
following isomorphisms of vector spaces:
\begin{equation}
\label{eq:isomorphic}
\Pi_n:M_1 \stackrel{\cong}{\longrightarrow} M_{n,1}
\quad\text{and}\quad
\pi_n\Phi_n:M_{n,1} \stackrel{\cong}{\longrightarrow} E_{n,1},
\end{equation}
where, strictly speaking, the isomorphisms are defined by the
restriction of $\Pi_n$ and $\pi_n\Phi_n$ to $M_1$ and $M_{n,1}$,
respectively.

The functions $g_{n,k}:=\Pi_n \mathbf 1_{\mathcal C_k}$,
$k=1,\ldots,\ell$ are in $M_{n,1}$ and converges to $\mathbf
1_{\mathcal C_k}$ in $W$-norm. Then, the vectors $\vartheta_{n,k}
=\pi_n(g_{n,k}\circ \varphi_n^{-1})$ are in $E_{n,1}$ and, as $n\to\infty$,
\begin{equation}
\label{eq:vartheta.converge}
\vartheta_{n,k,j}=\Pi_n(\mathbf 1_{\mathcal{C}_k})\circ\varphi_n^{-1}(X_j) 
\to
\mathbf 1_{\mathcal{C}_k}(X_j)=
\begin{cases}
1 & \text{if }k=k(j),\\
0 & \text{otherwise}.
\end{cases}
\end{equation}

Since $V_{n,1}$, \ldots, $V_{n,\ell}$ form a basis of $E_{n,1}$, there
exists a matrix $\xi_n$ of dimension $\ell\times\ell$ such that
\[
\vartheta_{n,k}=\sum_{i=1}^\ell \xi_{n,k,i} \, V_{n,i}.
\]
Hence the $j^\text{th}$ component of $\vartheta_{n,k}$, for all $j\in J(n)$,
may be expressed as
\[
\vartheta_{n,k,j}=\sum_{i=1}^\ell \xi_{n,k,i}\, V_{n,i,j}.
\]
Since $\rho_n(X_j)$ is the vector of $\mathbb R^\ell$ with components
$\{V_{n,i,j}\}_i$, the vector $\vartheta_{n,\bullet, j}
=\{\vartheta_{n,k,j}\}_k$ of $\mathbb R^\ell$
is related to $\rho_n(X_j)$ by the linear transformation $\xi_n$, i.e.,
\[
\vartheta_{n,\bullet, j}= \xi_{n}\, \rho_n(X_j).
\]
The convergence of $\vartheta_{n,\bullet, j}$ to $e_{k(j)}$ then
follows from \eqref{eq:vartheta.converge} and
Corollary~\ref{theo:strong} is proved. 
\\

\textbf{Acknowledgments.}  \textcolor{blue}{This work was supported by the French National Research Agency (ANR) under grant ANR-09-BLAN-0051-01.}

\bibliographystyle{abbrvnat}
\bibliography{spectralclustering.bib}


\begin{appendix}

\footnotesize

\section{Geometry of level sets}
The proof of the following result is adapted from Theorem~3.1 in
\cite{milnor} p.12 and Theorem~5.2.1 in 
\cite{jost} p.176.  
\begin{lem}
\label{lem:levelsets}
Let $f:\mathbb{R}^d\to\mathbb{R}$ be a function of class
$\mathcal{C}^2$.  Let $t\in\mathbb{R}$ and suppose that there exists
$\varepsilon_0>0$ such that
$f^{-1}\big([t-\varepsilon_0;t+\varepsilon_0]\big)$ is non empty,
compact and contains no critical point of $f$.  Let
$\{\varepsilon_n\}_n$ be a sequence of positive numbers such
that $\varepsilon_n<\varepsilon_0$ for all $n$, and $\varepsilon_n\to
0$ as $n\to\infty$.  Then there exists a sequence of diffeomorphisms
$\varphi_n:\mathcal{L}(t)\to\mathcal{L}(t-\varepsilon_n)$ carrying
$\mathcal{L}(t)$ to $\mathcal{L}(t-\varepsilon_n)$ such that:\\
1.  $\displaystyle\sup_{x\in\mathcal{L}(t)}\|\varphi_n(x)-x\|\to 0$ and
\\
2.  $\displaystyle\sup_{x\in\mathcal{L}(t)}\|D_x\varphi_n(x)-I_d\|\to 0$,
\\
as $n\to\infty$,
where $D_x\varphi_n$ denotes the differential of $\varphi_n$ and where $I_d$ is
the identity matrix on $\mathbb{R}^d$.
\end{lem}

\noindent{\bf Proof.}
Recall first that a one-parameter group of
diffeomorphisms $\{\varphi_u\}_{u\in\mathbb{R}}$ of $\mathbb{R}^d$ gives rise to a vector
field $V$ defined by
$$V_x g = \lim_{u\to 0}\frac{g\big(\varphi_u(x)\big) - g(x)}{u},\quad x\in\mathbb{R}^d,$$
for all smooth function $g:\mathbb R^d \to \mathbb R$.  Conversely, a
smooth vector field which vanishes outside of a compact set generates
a unique one-parameter group of diffeomorphisms of $\mathbb{R}^d$; see
Lemma~2.4 in \cite{milnor} p. 10 and Theorem 1.6.2 in \cite{jost}
p. 42.\\

Denote the set $\{x\in\mathbb{R}^d\,:\,a\leq f(x)\leq b\}$ by
$\mathcal{L}_a^b$, for $a\leq b$.  Let
$\eta:\mathbb{R}^d\to\mathbb{R}$ be the non-negative differentiable
function with compact support defined by
\[
\eta(x)=
\begin{cases}
1/\|D_xf(x)\|^2 
& \text{if $x\in\mathcal{L}_{t-\varepsilon_0}^t$},
\\
(t+\varepsilon_0-f(x))/
{\|D_xf(x)\|^2} & \text{if $x\in\mathcal{L}_t^{t+\varepsilon_0}$},
\\
0 & \text{otherwise}.
\end{cases}
\]
Then the vector field $V$ defined by $V_x = \eta(x)D_xf(x)$
has compact support $\mathcal{L}_{t-\varepsilon_0}^{t+\varepsilon_0}$,
so that $V$ generates a one-parameter group of diffeomorphisms
$$\varphi_u:\mathbb{R}^d\to\mathbb{R}^d,\quad u\in\mathbb R.$$
We have
\begin{equation*}
D_u\left[f\big(\varphi_u(x)\big)\right]
 =  \langle V , D_xf\rangle_{\varphi_u(x)}
  \geq  0,
\end{equation*}
since $\eta$ is non-negative.
Furthermore,
\[ 
\langle V , D_xf\rangle_{\varphi_u(x)} =1,
\quad \text{if }\varphi_u(x)\in\mathcal{L}_{t-\varepsilon_0}^t
\]
Consequently the map $u\mapsto f\big(\varphi_u(x)\big)$ has constant
derivative $1$ as long as $\varphi_u(x)$ lies in
$\mathcal{L}_{t-\varepsilon_0}^t$.  This proves the existence of the
diffeomorphism $\varphi_n:=\varphi_{-\varepsilon_n}$ which carries
$\mathcal{L}(t)$ to $\mathcal{L}(t-\varepsilon_n)$.\\

Note that the map $u\in\mathbb R\mapsto \varphi_u(x)$ is the integral curve
of $V$ with initial condition $x$.  Without loss of generality,
suppose that $\varepsilon_n\leq 1$.  For all $x$ in
$\mathcal{L}_{t-\varepsilon_0}^{t+\varepsilon_0}$, we have
\begin{equation*}
  \|\varphi_n(x)-x\|  \leq  
  \int_{-\varepsilon_n}^0 \left\|
    D_u\big(\varphi_u(x)\big)\right\|du
  \leq  \varepsilon_n/\beta(\varepsilon_n)\le
  \varepsilon_n/\beta(\varepsilon_0)
\end{equation*}
where we have set
\[
\beta(\varepsilon):=\inf\big\{\|D_xf(x)\|\,:\,
  x\in\mathcal{L}_{t-\varepsilon}^{t+\epsilon}\big\}>0.\]
This proves the statement 1, since $\varphi_n(x)-x$ is identically 0
on $\mathcal{L}(t+\varepsilon_0)$.\\

For the statement 2, observe that $\varphi_u(x)$ satisfies the relation
$$
\varphi_u(x) - x = 
\int_0^u D_v\big(\varphi_v(x)\big)dv 
= \int_0^u V\big(\varphi_v(x))\big)dv.
$$
Differentiating with respect to $x$ yields
$$
D_x\varphi_u(x) -I_d = \int_0^u D_x\varphi_v(x) 
\circ D_xV\big(\varphi_v(x)\big)dv.
$$
Since $f$ is of class $\mathcal{C}^2$, the two terms inside the
integral are uniformly bounded over
$\mathcal{L}_{t-\varepsilon_0}^{t+\varepsilon_0}$, so that there
exists a constant $C>0$ such that
$$\|D_x\varphi_n - I\|_x \leq C \varepsilon_n,$$
for all $x$ in $\mathcal{L}_{t-\varepsilon_0}^{t+\varepsilon_0}$.
Since $\|D_x\varphi_n - I\|_x$ is identically zero on
$\mathcal{L}(t+\varepsilon_0)$, this proves the statement 2.
\nopagebreak\hfill$\Box$

\section{Continuity of an isolated finite set of eigenvalues}
\label{appendix:B}

In brief, the spectrum $\sigma(T)$ of a
bounded linear operator $T$ on a Banach space is upper semi-continuous
in $T$, but not lower semi-continuous; see \cite{Kato}IV{\S}3.1 and
IV{\S}3.2. However, an isolated finite set of eigenvalues of $T$ is
continuous in $T$, as stated in Theorem~\ref{theo:continuous} below.

Let $T$ be a bounded operator on the $\mathbb C$-Banach space
$E$ with spectrum $\sigma(T)$.
Let
$\sigma_1(T)$ be a finite set of eigenvalues of $T$.
Set $\sigma_2(T)=\sigma(T)\setminus \sigma_1(T)$ and suppose that
$\sigma_1(T)$ is separated from 
$\sigma_2(T)$
by a
rectifiable, simple, and closed curve $\Gamma$.
Assume that a neighborhood
of $\sigma_1(T)$ is enclosed in the interior of $\Gamma$.  Then we have the
following theorem; see \cite{Kato}, III.{\S}6.4 and
III.{\S}6.5.
\begin{theo}[Separation of the spectrum]
  \label{theo:separation}
  The Banach space $E$ decomposes into a pair of
  supplementary subspaces as $E=M_1\oplus M_2$ such that 
  $T$ maps $M_j$ into $M_j$ ($j=1,2$) and
  the spectrum of the operator induced by $T$ on $M_j$ is
  $\sigma_j(T)$ ($j=1,2$).
  If additionally the total multiplicity $m$ of $\sigma_1(T)$
  is finite, then $\dim(M_1)=m$.
\end{theo}
Moreover, the following theorem states that a finite system of
eigenvalues of $T$, as well as the decomposition of $E$ of
Theorem~\ref{theo:separation}, depends continuously of $T$, see
\cite{Kato}, IV.{\S}3.5.  Let $\{T_n\}_n$ be a sequence of operators
which converges to $T$ in norm.  Denote by $\sigma_1(T_n)$ the part of
the spectrum of $T_n$ enclosed in the interior of the closed curve
$\Gamma$, and by $\sigma_2(T_n)$ the remainder of the spectrum of
$T_n$.
\begin{theo}[Continuous approximation of the spectral decomposition]
  \label{theo:continuous}
  There exists a finite integer $n_0$ such that the following holds
  true. 
  \\
  1. Both $\sigma_1(T_n)$ and $\sigma_2(T_n)$ are nonempty for all
  $n\ge n_0$ provided this is true for $T$.
  \\
  2. For each $n\ge0$, the Banach space $E$ decomposes into two
  subspaces as $E=M_{n,1}\oplus M_{n,2}$ in the manner of
  Theorem~\ref{theo:separation}, i.e.  $T_n$ maps $M_{n,j}$ into
  itself and the spectrum of $T_n$ on $M_{n,j}$ is
  $\sigma_j(T_n)$.
  \\
  3. For all $n\ge n_0$, $M_{n,j}$ is isomorphic to $M_j$.
  \\
  4. If $\sigma_1(T)$ is a singleton $\{\lambda\}$, then every sequence
  $\{\lambda_n\}_n$ with $\lambda_n\in\sigma_1(T_n)$ for all 
  $n\ge n_0$ converges to $\lambda$.
  \\
  5. If $\Pi$ is the projector on $M_1$ along $M_2$ and $\Pi_n$ the
  projector on $M_{n,1}$ along $M_{n,2}$, then $\Pi_n$ converges in
  norm to $\Pi$.
  \\
  6. If the total multiplicity $m$ of $\sigma_1(T)$ is finite, then,
  for all $n\ge n_0$, 
  the total multiplicity of $\sigma_1(T_n)$ is also $m$ and
  $\dim(M_{n,1})=m$.
\end{theo}

\section{Markov chains and limit operator}
\label{appendix:C}

For the reader not
familiar with Markov chains on a general state space, we begin by
summarizing the relevant
part of the theory.

\subsection{Background materials on Markov chains}
\label{sub:background}
Let $\{\xi_i\}_{i\ge0}$ be a Markov chain with state space $\mathcal
S\subset \mathbb R^d$ and transition kernel $q(x,dy)$.  We write $P_x$
for the probability measure when the initial state is $x$ and $E_x$
for the expectation with respect to $P_x$.  The Markov chain is called
\textit{(strongly) Feller} if the map
\[
x\in \mathcal S \mapsto Qg(x):=\int_{\mathcal
  S}q(x,dy)g(y)=\mathbb{E}_xf(\xi_1)
\]
is continuous for every bounded, measurable function $g$ on $\mathcal
S$; see \cite{MT}, p. 132. This condition ensures that the chain
behaves nicely with the topology of the state space $\mathcal S$.  The
notion of irreducibility expresses the idea that, from an arbitrary
initial point, each subset of the state space may be reached by the
Markov chain with a positive probability. A Feller chain is said
\textit{open set irreducible} if, for every points $x, y$ in $\mathcal
S$, and every $\eta>0$,
\[
\sum_{n\ge 1} q^n(x, y+\eta B)>0,
\]
where $q^n(x,dy)$ stands for the $n$-step transition kernel;
see \cite{MT}, p. 135.  Even if open set irreducible, a Markov chain
may exhibit a periodic behavior, i.e., there may exist a
partition $\mathcal S=\mathcal S_0 \cup \mathcal S_1\cup \ldots \cup
\mathcal S_N$ of the state space such that, for every initial state
$x\in \mathcal S_0$,
\[
P_x[\xi_1\in \mathcal S_1, \xi_2\in \mathcal S_2, 
\ldots, \xi_N\in \mathcal S_N, \xi_{N+1}\in \mathcal S_0,
\ldots] =1.
\]
Such a behavior does not occur if the Feller chain is \textit{topologically aperiodic}, i.e., if for each
initial state $x$, each $\eta>0$, there exists $n_0$ such that
$q^n(x,x+\eta
B)>0$ for every $n\ge n_0$; see \cite{MT}, p. 479.\\

Next we come to ergodic properties of the Markov chain.
 A Borel set
$A$ of $\mathcal S$ is called \textit{Harris recurrent} if the chain
visits $A$ infinitely often with probability 1 when started at any
point $x$ of $A$, i.e.,
\[
P_x\left(
\sum_{i=0}^\infty \mathbf 1_A(\xi_i)=\infty
\right)=1
\]
for all $x\in A$.
The chain is then said to be \textit{Harris
  recurrent} if every Borel set $A$ with positive Lebesgue measure is
Harris recurrent; see \cite{MT}, p. 204.
At least two types of behavior, called evanescence and
non-evanescence, may occur.
The event $[\xi_n\to\infty]$ denotes the fact that
the sample path visits each compact set only finitely many often, and
the Markov chain is called \textit{non-evanescent} if $P_x(\xi_n\to
\infty)=0$ for each initial state $x\in \mathcal S$. Specifically,
a Feller chain is Harris recurrent if and only if it
is non-evanescent; see \cite{MT}, Theorem 9.2.2, p. 212.\\

The ergodic properties exposed above describe the long time behavior of the chain. A
measure $\nu$ on the state space is said \textit{invariant} if
\[
\nu(A)=\int_{\mathcal S}q(x,A)\nu(dx)
\]
for every Borel set $A$ in $\mathcal S$.  If the chain is Feller, open
set irreducible, topologically aperiodic and Harris recurrent, it
admits a unique (up to constant multiples) invariant measure $\nu$;
see \cite{MT}, Theorem 10.0.1 p. 235.
In this case, either
$\nu(\mathcal S)<\infty$ and the chain is called \textit{positive}, or
$\nu(\mathcal S)=\infty$ and the chain is called \textit{null}.
The following important result provides one with the limit of the distribution of $\xi_n$ when
$n\to\infty$, whatever the initial state is.
Assuming that
the chain is Feller, open set irreducible, topologically aperiodic and
positive Harris recurrent, the sequence of distribution
$\{q^n(x,dy)\}_{n\ge 1}$ converges in total variation to
$\nu(dy)$, the unique invariant probability distribution; see Theorem
13.3.1 of \cite{MT}, p. 326. That is to say, for every $x$ in
$\mathcal S$,
\[
\sup_g\left\{
  \left|\int_{\mathcal S} g(y) q^n(x,dy)-\int_{\mathcal S}
    g(y)\nu(dy)\right|
\right\} \to 0
\quad \text{as } n\to\infty,
\]
where the supremum is taken over all continuous functions $g$ from
$\mathcal S$ to $\mathbb R$ with $\|g\|_\infty\le 1$.

\subsection{Limit properties of $Q_h$}
With the definitions and results from the previous paragraph, we may
now study the properties of the limit clustering induced by the
operator $Q_h$.  The transition kernel $q_h(x,dy):=q_h(x,y)\mu^t(dy)$
defines a Markov chain with state space $\mathcal L(t)$.  Recall 
that $\mathcal{L}(t)$ has $\ell$ connected components
$\mathcal{C}_1,\dots,\mathcal{C}_\ell$ and that under Assumption~3,
$h$ is strictly lower than $d_{min}$, the minimal distance between the
connected components.

\begin{pro}
  \label{prop:Harris}
    1. The chain is Feller and topologically aperiodic.
      \\
    2. When started at a point $x$ in some connected component of
      the state space, the chain evolves within this connected
      component only.    
      \\
    3. When the state space is reduced to some connected component
      of $\mathcal L(t)$, the chain is open set irreducible
      and positive Harris recurrent.
\end{pro}

\noindent\textbf{Proof.}
  1. Since the similarity function $k_h$ is continuous, with compact
  support $hB$, the map
  \[
  x\mapsto Q_hg(x)=\int_{\mathcal L(t)} q_h(x,dy)g(y)
  \]
  is continuous for every bounded, measurable function $g$. Moreover,
  $k_h$ is bounded from below on $(h/2)B$ by Assumption~2. Thus, for each
  $x\in\mathcal L(t)$, $n\ge1$ and $\eta>0$,
  $q_h^n(x,x+\eta B)>0$. Hence, the chain is Feller and
  topologically aperiodic.\\
  
  2. Without loss of generality, assume that $x\in\mathcal
  C_1$.
  Let $y$ be a point of $\mathcal L(t)$ which does not belong to $\mathcal C_1$.
  Then $\|y-x\|\ge d_{min}>h$ so that $q_h(x,y)=0$.
  Whence,
  \[
  P_x(\xi_1\in \mathcal C_1) = q_h(x,\mathcal C_1) = \int_{\mathcal
    C_1} q_h(x,y) \mu^t(dy) = \int_{\mathcal L(t)} q_h(x,y)
  \mu^t(dy)=1.
  \]
\medskip

  3. Assume that the state space is reduced to $\mathcal C_1$.  Fix
  $x, y\in \mathcal C_1$ and $\eta>0$.
  Since $\mathcal{C}_1$ is connected, there exists a finite
  sequence $x_0$, $x_1$, \ldots $x_N$ of points in $\mathcal C_1$ such
  that $x_0=x$, $x_N=y$, and $\|x_i-x_{i+1}\|\le h/2$ for each
  $i$. Therefore
  \[
  q_h^N(x,y+\eta B) \ge P_x(\xi_i\in x_i + \eta B 
  \text{ for all  }i\le N) >0
  \]
  which proves that the chain is topologically aperiodic.

  Since $\mathcal C_1$ is compact, the chain is non-evanescent, and so
  it is Harris recurrent. Recall that $k(x)=k(-x)$ from Assumption
  2. Therefore $k_h(y-x)=k_h(x-y)$ which yields 
  \[
  K_h(x)q_h(x,dy)\mu^t(dx)=K_h(y)q_h(y,dx)\mu^t(dy).
  \]
  By integrating the previous relation with respect to $x$ over $\mathcal C_1$, one may verify
  that $K_h(x)\mu^t(dx)$ is an invariant measure.
  At last $\int_{\mathcal
    C_1} K_h(x) \mu^t(dx)<\infty$, which proves that the chain is
  positive.
\nopagebreak\hfill$\Box$


\begin{pro}
\label{pro:harmonic}
If $g$ is continuous and $Q_hg=g$, then
$g$ is constant on the connected components of $\mathcal L(t)$.
\end{pro}
\noindent\textbf{Proof.}
  We will prove that $g$ is constant over $\mathcal
  C_1$.
  Proposition~\ref{prop:Harris} provides one with a unique invariant
  measure $\nu_1(dy)$ when the state space is reduced to $\mathcal
  C_1$. Fix $x$ in $\mathcal C_1$.  Since $g=Q_hg$, $g=Q_h^ng$ for
  every $n\ge 1$.
  Moreover
  by Proposition~\ref{prop:Harris}, the chain is open set
  irreducible, topologically aperiodic, and positive Harris recurrent
  on $\mathcal C_1$. Thus, $q_h^n(x,dy)$ converges in total variation
  norm to $\nu_1(dy)$. Specifically,
  \[
  Q_h^ng(x) \longrightarrow \int_{\mathcal C_1}
  g(y)\nu_1(dy) \quad \text{as } n\to\infty.
  \]
  Hence, for every $x$ in $\mathcal C_1$, 
  \[
  g(x)=\int_{\mathcal C_1} g(y)\nu_1(dy),
  \]
  and since the last integral does not depend on $x$, it follows that
  $g$ is a constant 
  function on $\mathcal{C}_1$.  \nopagebreak\hfill$\Box$

\end{appendix}



\end{document}